\documentclass[nohyperref]{article}

\usepackage{microtype}
\usepackage{graphicx}
\usepackage{subfigure}
\usepackage{booktabs} 

\usepackage{hyperref}

\usepackage[accepted]{icml2022}

\usepackage{amsmath}
\usepackage{amssymb}
\usepackage{mathtools}
\usepackage{amsthm}
\usepackage{bm}
\usepackage[mathscr]{eucal}
\usepackage{caption}

\usepackage[capitalize,noabbrev]{cleveref}
\theoremstyle{plain}
\newtheorem{theorem}{Theorem}[section]

\theoremstyle{definition}
\newtheorem{definition}[theorem]{Definition}

\theoremstyle{remark}

\usepackage[textsize=tiny]{todonotes}

\icmltitlerunning{History Compression via Language Models in Reinforcement Learning}

\usepackage{array}

\newcommand{\eg}{e.\,g., }
\newcommand{\ie}{i.\,e., }

\newcommand{\ourmethod}{HELM} 
\newcommand{\frozenpool}{FrozenHopfield}
\newcommand{\baseline}{Impala-PPO}
\newcommand{\doorkey}{DoorKey}
\newcommand{\redblue}{RedBlueDoors}
\newcommand{\dynobst}{DynamicObstacles}
\newcommand{\keycorr}{KeyCorridor}

\newcommand\Ba{\bm{a}}

\newcommand\Bd{\bm{d}}
\newcommand\Be{\bm{e}}

\newcommand\Bh{\bm{h}}

\newcommand\Bo{\bm{o}}

\newcommand\Bs{\bm{s}}

\newcommand\Bx{\bm{x}}

\newcommand\Bz{\bm{z}}
\newcommand\BE{\bm{E}}

\newcommand\BP{\bm{P}}
\newcommand\BQ{\bm{Q}}
\newcommand\BR{\bm{R}}

\newcommand\Bxi{\bm{\xi}}

\newcommand{\dE}{\mathbb{E}}

 \newcommand{\dN}{\mathbb{N}}
 \newcommand{\dP}{\mathbb{P}} 
 \newcommand{\dR}{\mathbb{R}}

 \newcommand{\cB}{\mathcal{B}}

 \newcommand{\cN}{\mathcal{N}}

\newcommand{\norm}[1]{{{\left\|#1\right\|}}} 

\makeatletter
\newcommand{\dlmf}[1]{%
\citep[%
  \def\nextitem{\def\nextitem{, }}%
  \@for \el:=#1\do{\nextitem\href{http://dlmf.nist.gov/\el}{(\el)}}%
]{olver_nist_2010}%
}
\makeatother

\begin{document}

\twocolumn[
\icmltitle{History Compression via Language Models in Reinforcement Learning}



\icmlsetsymbol{equal}{*}

\begin{icmlauthorlist}
\icmlauthor{Fabian Paischer}{jku,ellis,equal}
\icmlauthor{Thomas Adler}{jku,equal}
\icmlauthor{Vihang Patil}{jku}
\icmlauthor{Angela Bitto-Nemling}{jku,ellis,iarai}
\icmlauthor{Markus Holzleitner}{jku}
\icmlauthor{Sebastian Lehner}{jku,ellis}
\icmlauthor{Hamid Eghbal-zadeh}{jku}
\icmlauthor{Sepp Hochreiter}{jku,ellis,iarai}
\end{icmlauthorlist}

\icmlaffiliation{jku}{LIT AI Lab, Institute for Machine Learning, Johannes Kepler University Linz, Austria}
\icmlaffiliation{ellis}{ELLIS Unit Linz}
\icmlaffiliation{iarai}{Institute of Advanced Research in Artificial Intelligence (IARAI), Vienna, Austria}

\icmlcorrespondingauthor{Fabian Paischer}{paischer@ml.jku.at}

\icmlkeywords{Machine Learning, ICML}

\vskip 0.3in
]



\printAffiliationsAndNotice{\icmlEqualContribution}  

\begin{abstract}
In a partially observable Markov decision process (POMDP), an agent typically  
uses a representation of the past to approximate the underlying MDP.
We propose to utilize a frozen Pretrained Language Transformer (PLT)
for history representation and compression 
to improve sample efficiency.
To avoid training of the Transformer, we introduce \textit{\frozenpool},
which automatically associates observations with pretrained token embeddings.
To form these associations,
a modern Hopfield network stores these token embeddings,
which are retrieved by queries that are obtained by a
random but fixed projection of observations.
Our new method, \ourmethod, enables actor-critic network architectures that contain  
a pretrained language Transformer
for history representation as a memory module.
Since a representation of the past need not be learned,
\ourmethod~is much more sample efficient than competitors.
On Minigrid and Procgen environments \ourmethod~achieves new state-of-the-art results. 
Our code is available at \url{https://github.com/ml-jku/helm}.
\end{abstract}

\section{Introduction}

Recently, reinforcement learning (RL) for partially observable Markov decision processes (POMDPs; \citealp{astrom_optimal_1964,kaelbling_planning_1998})
became popular and attracted a lot of attention 
through mastering complex games such as DotA \citep{berner_dota_2019} or Starcraft \citep{vinyals_grandmaster_2019}. 
In POMDPs, only observations are available but the state of the underlying Markov decision process (MDP) 
is not known. 
To approximate the state most methods use the current observation together with a memory mechanism, 
such as an LSTM \citep{hochreiter_long_1997}, that stores past observations.
The memory stores only abstractions of the raw observations to allow for generalization and effective processing.
Abstractions are also used in human language when describing 
objects, their mutual relations, events, situations, concepts and ideas.
Human language is tailored to represent and compress the past via abstraction
to convey past experiences and events from one generation to the next.
History compression only stores observations that are relevant, \ie
contain new information \cite{schmidhuber_learning_1992}.
Abstraction is a form of compression that 
maps segments of constituent information to a single segment of information \citep{chaitin_limits_2006,gell-mann_what_1995}. 
During evolution, human language has been optimized with respect to compression and abstraction for being effective in storing past experiences of one human to pass them on to other humans \citep{nowak_evolution_1999}. 
Thus, language is inherently well-suited for constructing abstractions and representing the past.
Language models are usually scaled up and pretrained on large-scale text corpora. 
During pretraining, language models 
learn to compress information to represent abstract concepts prevalent in natural language. 
This motivates us to leverage pretrained language models for history compression.

The Transformer \citep{vaswani_attention_2017} proved to be very effective for language modeling \citep{brown_language_2020,dai_transformer-xl_2019}.
Further, it has been successfully applied for sequence modeling in the domain of RL \citep{janner_reinforcement_2021,chen_decision_2021}.
Therefore, it is very well suited to represent and store past observations. 
The gated TransformerXL \citep{parisotto_stabilizing_2020} is a Transformer architecture 
tailored to on-policy reinforcement learning and trained from scratch.
However, it tends to overfit on 
the usually scarce data generated by the current policy.
The same problem appears when finetuning a large pretrained Transformer model.
To avoid these overfitting problems,
we use a Transformer pretrained on a large text corpora without finetuning or
weight adjustments.
To enable the usage of a fixed transformer, we propose our new \textit{\frozenpool} that maps
observations and actions to the input space of the Transformer, without any training, by associating them with token embeddings.
These associations are obtained by first randomly projecting observations and actions
to state patterns (queries) and then using them to retrieve token embeddings 
from a modern Hopfield network (MHN, \citealp{ramsauer_hopfield_2021}).
The MHN stores the pretrained token embeddings, therefore their
retrievals are in the simplex spanned by these token embeddings.
We propose \ourmethod~(History comprEssion via Language Models)
for learning a policy by any RL algorithm like, e.g., Proximal Policy Optimization 
(PPO; \citealp{schulman_proximal_2017}) in POMDPs.
\ourmethod~uses novel policy and value network architectures 
that include a pretrained Transformer
for abstract history representation and compression.
If the current state
can be predicted from the current observation and the history,
then \ourmethod~even transforms the POMDP into an MDP.
Since \ourmethod~
does not require learning a history representation,
it is much more sample efficient than previous methods. 

Our contributions are as follows: 
\begin{itemize}
\item We introduce \textit{\frozenpool}, which
automatically associates actions and observations with original token embeddings of the Transformer
without training or weight adjustments while still being theoretically sound.
\item We introduce \ourmethod, a framework for actor-critic network architectures with 
a pretrained language Transformer
for abstract history representation as a memory module.
\end{itemize}

\section{History Compression via Language Models}
\label{sec:methodology}

Our target is to leverage a Pretrained Language Transformer without any finetuning in the highly sensitive setting of on-policy RL for partially observable environments. This raises two problems: (i) the problem of finding a suitable translation from environment observations to the language domain, and (ii) how to effectively utilize the compressed history for policy optimization. To address (i), we introduce \textit{\frozenpool}, a frozen associative memory based on random projections as queries that map from any modality other than language to a token representation. We feed sequences of projected environment observations into the PLT and retrieve a compressed representation of the current episode history. To tackle (ii), we propose a simple framework in which the representation of the past is concatenated with a learned encoding of the current observation via a convolutional neural network (CNN; \citealp{fukushima_neocognitron_1980,lecun_backpropagation_1989}), and serves as input to the policy optimization algorithm. The actor-critic networks consist of shallow multi-layer perceptrons which are, along with the CNN encoder, the only trainable components in our proposed setting. Thus, we leverage the ability to abstract information learned during pretraining of PLTs for partially observable environments in RL. 

\begin{figure}[!ht]
\vskip 0.2in
\begin{center}
\centerline{\includegraphics[width=\columnwidth]{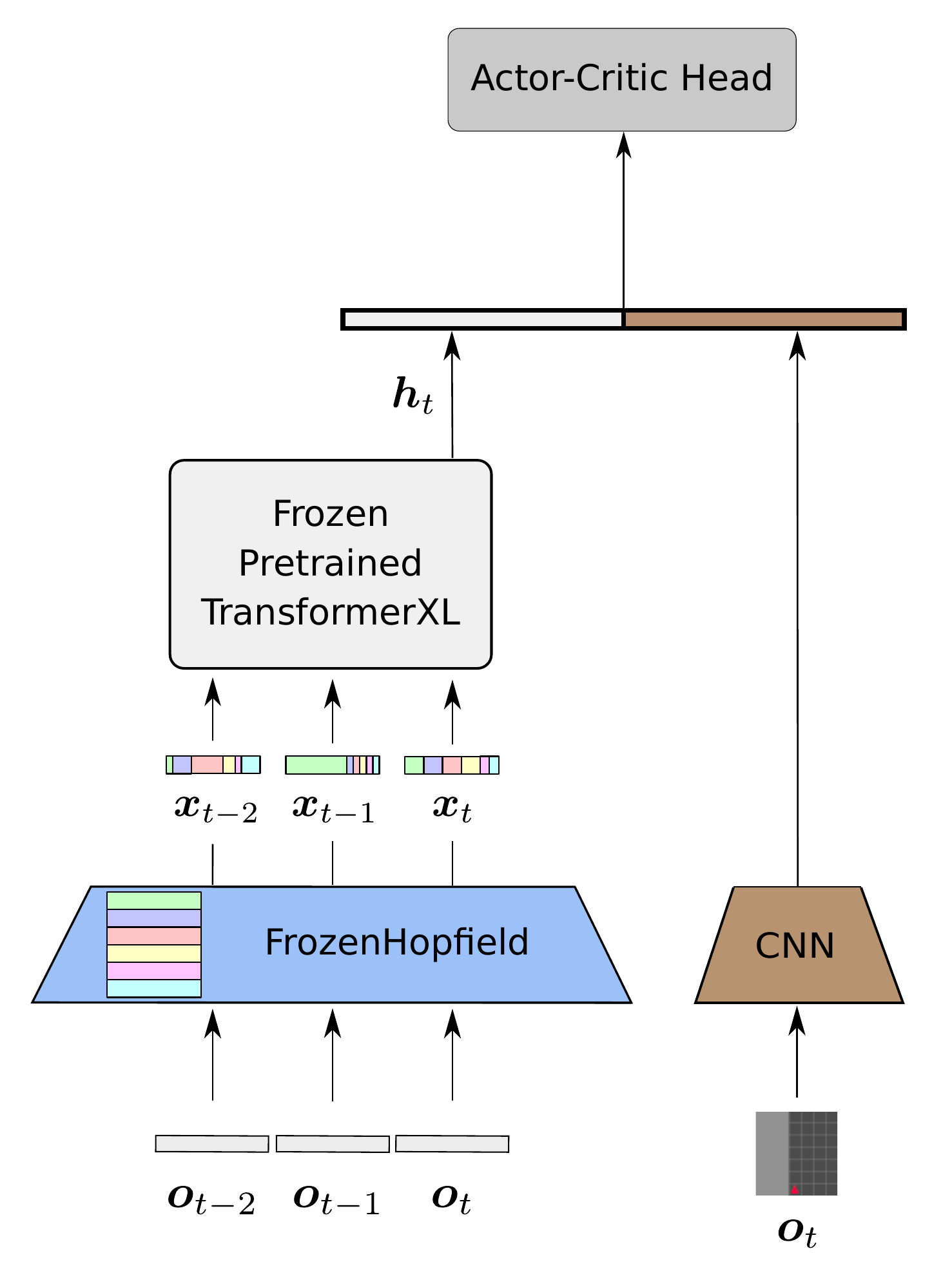}}
\caption{Architecture of \ourmethod: The sequence of observations is encoded via the \textit{\frozenpool} mechanism and fed into the TransformerXL. The current timestep is encoded with a CNN and concatenated with the output of the language model to form the input for the actor-critic head.}\label{fig:methodology}
\end{center}
\vskip -0.2in
\end{figure}

\begin{figure}[!ht]
\vskip 0.2in
\begin{center}
\centerline{\includegraphics[width=\columnwidth]{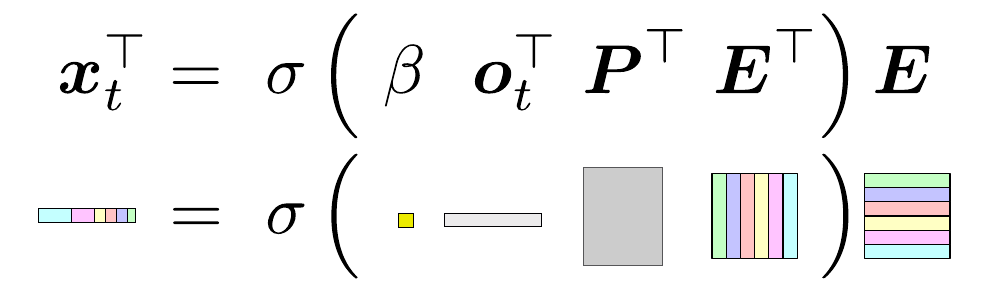}}
\caption{The \textit{\frozenpool} mechanism is realized as a random but fixed retrieval from a MHN. The current observation $\Bo_t$ forms the query and the pretrained token embeddings of the language model $\BE$ are used as stored patterns. The random projection $\BP$ is leveraged for projecting the observations to the same dimensionality as the token embeddings. $\beta$ is the inverse temperature that determines how close the retrievals are to the stored tokens. The retrieval $\Bx_t$ is located in the simplex spanned by the rows of $\BE$.} \label{fig:FrozenHopfield}
\end{center}
\vskip -0.2in
\end{figure}

Our proposed framework, \ourmethod, consists of three main parts: (i) the memory mechanism which utilizes a pretrained language model, (ii) an encoder of the current timestep, and (iii) the actor-critic networks trained for policy optimization. 
We instantiate (i) with a TransformerXL (TrXL;~\citealp{dai_transformer-xl_2019}).
For (ii) we utilize a CNN based on the small Impala architecture \citep{espeholt_impala_2018} since we mainly focus on image-based environments. Finally, we utilize a MLP with one hidden layer for (iii). \cref{fig:methodology} shows a graphical illustration of \ourmethod. Pseudocode can be found in \cref{fig:alg1}.

We use a TrXL that is pretrained via Causal Language Modeling (CLM) based on next-token prediction. 
Thus, during pretraining the information of an entire trajectory is compressed in the most recent token, based on which the next token is predicted. 
TrXL uses relative positional encodings and provides a memory register which stores past activations and allows processing sequences beyond a fixed context length. 
This property is beneficial in the context of RL where the length of each episode is unknown a-priori during inference. 
Also leveraging the memory register drastically reduces computational complexity. 
Each time a new observation is processed, its corresponding activations are shifted into the memory register, which allows feeding a single observation at a time into the model while still preserving the context up to a certain length.
We use the huggingface implementation \citep{wolf_transformers_2020} of TrXL consisting of 257M parameters (18 hidden layers, 1024 hidden units per layer, 16 attention heads), which is pretrained on the WikiText 103 dataset \citep{merity_pointer_2017} . 
We found TrXL to strike a reasonable balance between complexity and performance.

In order to utilize the Transformer for RL we aim to find a suitable input representation for each environment observation in the Transformer embedding space. 
Finding an efficient mapping from, \eg images to the language domain is not straightforward and usually involves training. 
A common approach is to learn an input projection either via a linear layer \citep{lu_pretrained_2021} or a convolutional neural network \citep{tsimpoukelli_multimodal_2021}. 
However this requires back-propagating gradients through the entire language model which is computationally expensive. 
To this end, we introduce \textit{\frozenpool}, a random but fixed Hopfield retrieval mechanism, which allows mapping inputs from any modality to the language domain by utilization of pretrained token embeddings. 
The \textit{\frozenpool} mechanism consists of two essential steps:

\begin{enumerate}
    \item Random projections of observations (and possibly actions) to the Transformer token embedding space
    \item Retrievals from the MHN that stores the original token embeddings and uses the projections as state patterns or queries 
\end{enumerate}

The first step is justified by the Johnson-Lindenstrauss (JL) lemma \citep{johnson_extensions_1984}, which guarantees approximate preservation of the mutual distances of the observations after a random projection with high probability. 
The second step ensures that after retrieval each observation is represented within the convex hull of the Transformer embeddings.
In combination these two steps associate observations from the environment with embeddings from the language domain of the pretrained Transformer without any training. 
We found that this mechanism provides a cheap and effective way to map observations to a suitable input representation for the Transformer. It even yields significantly better results in terms of sample efficiency as opposed to learning an input projection (see \cref{app:ablation}). \cref{fig:FrozenHopfield} shows a graphical illustration of the \textit{\frozenpool} mechanism, which we describe more formally in the following.

Let $\BE = (\Be_1, \dots, \Be_k)^\top \in \dR^{k \times m}$ 
be the token embedding matrix 
from the pretrained Transformer consisting of $k$ embeddings $\Be_i \in \dR^m$. 
At time $t$, we obtain inputs $\Bx_t \in \dR^m$ for the Transformer 
from observations $\Bo_t \in \dR^n$ via the \textit{\frozenpool} mechanism by 
\begin{equation}
    \Bx_t^\top = \sigma( \beta \Bo_t^\top \BP^\top \BE^\top) \BE, \label{eqn:frozen_pool}
\end{equation}
where $\sigma$ is the softmax function.
Further, $\BP \in \dR^{m \times n}$ is a random projection matrix whose entries are independently 
sampled from a $\cN(0, n / m)$ Gaussian distribution. 

The variance $n / m$ is a consequence of the JL lemma, which implies the following. 
Given two observations $\Bo_t, \Bo'_t$, let $\Bd = \Bo_t - \Bo'_t$.
For any $0 < \varepsilon < 1$ with probability at least $1 - \delta$ we have 
\begin{equation}
    (1-\varepsilon) \norm{\Bd}_2^2 \leq \norm{\BP \Bd}_2^2 \leq (1+\varepsilon) \norm{\Bd}_2^2,
    \label{eqn:jl_approx}
\end{equation}
where $\norm{\cdot}_2$ is the Euclidean norm and 
\begin{equation}
    \delta = 2 \exp{\left(-\frac{m (\varepsilon^2/2 - \varepsilon^3/3)}{2}\right)}.
    \label{eqn:jl_prob}
\end{equation}
This means that the projection $\BP$ is approximately distance preserving with high probability. Details can be found in \cref{app:jl_lemma}.

Equation~\eqref{eqn:frozen_pool}
is the Hopfield retrieval \citep{ramsauer_hopfield_2021}, 
where $\BP \Bo_t$ are the state patterns or queries 
and $\BE$ are the stored patterns. 
Equivalently, it can be viewed as attention 
mechanism with query $\BP \Bo_t$, and keys and values $\BE$.
See \cref{app:hopfield} for details.
Since the inputs $\Bx_t$ are in the simplex  
spanned by the stored patterns $\BE$, it is guaranteed that 
$\Bx_t$ lies in the convex hull of the set $\{\Be_1, \dots, \Be_k\}$. 
Thus, we feed the transformer with inputs from the regime it was trained on. 
The inverse temperature $\beta > 0$ of the MHN is a hyperparameter 
of our method and controls how close the retrievals are to the original token embeddings. 
For $\beta \to 0$, the mean $\frac1k \sum_{i=1}^k \Be_i$ is retrieved, independent of the query. 
Whereas for $\beta \to \infty$, the original token embedding $\Be_i$ is retrieved that is closest 
to the mapped $\Bo_t$, that is, $i = \arg \max_{i \in \{1, \dots, k\}} \Be_i^\top \BP \Bo_t$.

The representation of the current timestep is obtained via the CNN encoder. We specify the output of the encoder to be of the same dimensionality as the output of TrXL. The representation of the current timestep and the context representation $\Bh_t$ are concatenated and serve as input to the actor-critic head.
In order to enhance the reliance of the policy algorithm on the Transformer output, we construct the actor-critic networks as shallow as possible with only a single hidden layer. 
In principle, any algorithm for policy optimization may be used. 
In our experiments, we focus on PPO with multistep experience, clipped importance sampling ratio \citep{schulman_proximal_2017}, and generalized advantage estimation \citep{schulman_high-dimensional_2016}. We use the \texttt{stable-baselines3} \citep{raffin_stable_2019} package for the implementation of our algorithm.

\begin{figure}[ht]
\begin{algorithm}[H]
   \caption{HELM}
    \begin{algorithmic}
       \REQUIRE Interaction budget $N$, Horizon $T$, Actor-critic head $\pi$, Environment \texttt{env}, RL algorithm \texttt{update}
       \STATE $\cB \gets \emptyset$ \COMMENT{Initialize buffer}
       
       \FOR{$i=1$ {\bfseries to} $N$}
       \STATE $\Bo_1 \gets \texttt{env.reset}()$
       \FOR{$t=1$ {\bfseries to} $T$}
       \STATE $\Bx_{t} \gets \text{\frozenpool}(\Bo_{t})$ \COMMENT{Retrieve associations}
       \STATE $\Bh_{t} \gets \text{TrXL}(\Bx_{t})$ \COMMENT{Compress history}
       \STATE $\Tilde{\Bs}_{t} \gets [\Bh_{t};\text{CNN}(\Bo_{t})]$ \COMMENT{Concatenate}
       \STATE $\Ba_{t} \sim \pi(\Tilde{\Bs}_{t})$ \COMMENT{Sample action}
       \STATE $\Bo_{t+1}, r_{t+1} \gets \texttt{env.step}(\Ba_{t})$ \COMMENT{Interact}
       \STATE $\cB \gets [\cB; (\Bo_{t}, \Bh_{t}, \Ba_{t}, r_{t+1})]$ \COMMENT{Extend buffer}
       \ENDFOR
       \STATE $\texttt{update}(\text{CNN}, \pi, \cB)$
       \ENDFOR
    \end{algorithmic}
\end{algorithm}
\label{fig:alg1}
\setcounter{algorithm}{0}
\captionof{algorithm}{History Compression via \ourmethod. Observations are mapped to the language domain and processed by TrXL to obtain a compressed version of the history. Actor-critic networks are trained on the history representations concatenated with an encoding of the current timestep. TrXL can be exchanged with any other pretrained language model, and actor-critic are initialized depending on the RL algorithm. For on-policy RL the buffer $\cB$ corresponds to a rollout buffer, while for off-policy RL it is initialized as replay buffer \citep{lin_self-improving_1992}. The CNN encoder may be exchanged with any suitable encoder for the current observation.}
\end{figure}

\section{Experiments}
\label{sec:experiments}
We conduct experiments on partially observable environments to exploit the compressed history abstraction of the PLT. 
Furthermore, we train on procedurally generated environments which enhance diversity and force the agent to learn generalizable skills by sampling level configurations from a predefined distribution. 
As toytasks we generate a RandomMaze environment \citep{zuo_mazelab_2018}, select a memory-dependent environment from Minigrid (\keycorr, \citealp{chevalier-boisvert_minimalistic_2018}), and evaluate our agents on complex environments from the Procgen suite \citep{cobbe_leveraging_2020}. 

\ourmethod\ is compared to a recurrent baseline trained from scratch and evaluated for sample efficiency. 
As baseline we take the small Impala architecture (\baseline, \citealp{espeholt_impala_2018}), which consists of a convolutional backbone and an LSTM, since it reaches state-of-the-art performance on Procgen environments. 
To measure the dependence on memory we also include the performance of a Markovian baseline (CNN-PPO) which consists of the same architecture as \baseline\ without the LSTM. 
In accordance with recent work  \citep{jiang_prioritized_2021,goyal_recurrent_2021,madan_fast_2021}, we use PPO as on-policy algorithm for our Minigrid experiments.
For Procgen we show performance for PPO as in \cite{cobbe_leveraging_2020}.
Additionally, we provide results for the recently proposed phasic policy gradient (PPG) algorithm \citep{cobbe_phasic_2021} for all methods on the 6 Procgen environments in \cref{app:ppo_ppg}.
We evaluate for sample efficiency by measuring the performance at the end of training and test for statistical significance via a one-sided Wilcoxon rank-sum test \citep{wilcoxon_individual_1945} at a confidence level of $\alpha = 0.05\,\%$. 
The performance is evaluated by measuring the interquartile mean (IQM) and 95\,\% bootstrapped confidence intervals (CIs), as proposed in \citet{agarwal_deep_2021}.

The best hyperparameters for each method are determined via gridsearches, while others were set to fixed values to trade off speed and complexity. 
For the TrXL in \ourmethod, we choose a memory register length of 512 which includes the current timestep. 
The input to the \textit{\frozenpool}~component consists of flattened grayscaled  observations. 
Since there is no need to backpropagate through the Transformer, only its hidden representations for each timestep are stored in the rollout buffer. 
This alleviates the need to process entire trajectories for computing the gradients. 
The CNN encoder of \ourmethod, as well as CNN-PPO and \baseline\ are trained on RGB images. A hyperparameter search is performed for all methods over a small grid (see \cref{app:hyperparam}).

\subsection{RandomMaze}
RandomMaze provides a testbed for evaluating the memory capabilities of our methods. The agent is represented as a tile in a maze of randomly sampled size and receives a $9 \times 9$ agent centered excerpt of the state to enforce partial observability. A reward of -1 is elicited when bumping against a wall, which ends the current episode. 
Furthermore, at each interaction step a small negative reward of -0.01 is given. If the agent reaches the goal tile, the episode ends and a positive reward of 1 is elicited. More details on the RandomMaze environment can be found in \cref{app:randommaze}. 

To illustrate the dependence on memory we additionally train a Markovian baseline (CNN-PPO) which features the same CNN encoder as \ourmethod, and the actor-critic head, but no memory mechanism to store past events. \cref{tab:randommaze} shows the performance of our models. We train for 2M interaction steps and evaluate for sample efficiency at the end of training. The CNN-PPO agent performs surprisingly well under the partial observability in RandomMaze. 
\ourmethod~achieves the highest score indicating its advantage over CNN-PPO due to its history compression component. We find that \baseline~reaches a significantly lower score ($p=1\text{e-}4$) than \ourmethod\ which highlights the superior sample efficiency of \ourmethod.
This experiment also demonstrates that raw observations may be of greater or smaller dimensionality than the hidden dimension of TrXL before being processed with the \frozenpool~component.

\begin{table}[t]
\caption{IQM and 95\,\% bootstrapped CIs of the return over the last 100 episodes across 30 seeds on RandomMaze after 2M steps. Bold indicates maximum reached return.}
\label{tab:randommaze}
\vskip 0.15in
\begin{center}
\begin{small}
\begin{sc}
\begin{tabular}{lcc}
\toprule
\ourmethod & \baseline & CNN-PPO \\
\midrule
\textbf{0.185} $\pm$ 0.04 & 0.05 $\pm$ 0.06 & 0.16 $\pm$ 0.035 \\
\bottomrule
\end{tabular}
\end{sc}
\end{small}
\end{center}
\vskip -0.25in
\end{table}

\subsection{Minigrid}
Minigrid environments provide a variety of tasks which are designed to expose weaknesses of current RL algorithms. A sparse reward of 1 is elicited after the agent successfully completes the given task. For each interaction step a small value is subtracted from the reward, thus, the optimal score is close to 1. Furthermore, Minigrid environments are turned into POMDPs by restricting the field of view of the agent. In the \keycorr~environment the agent must first navigate to the left to pick up a key, and then navigate back to the right to unlock a door and pick up an object. In order to solve the task the agent must effectively represent various entities in its hidden representation and remember that it has picked up a key prior to an attempt of opening the door.

\begin{figure}[ht]
\vskip 0.2in
\begin{center}
\centerline{\includegraphics[width=.5\textwidth]{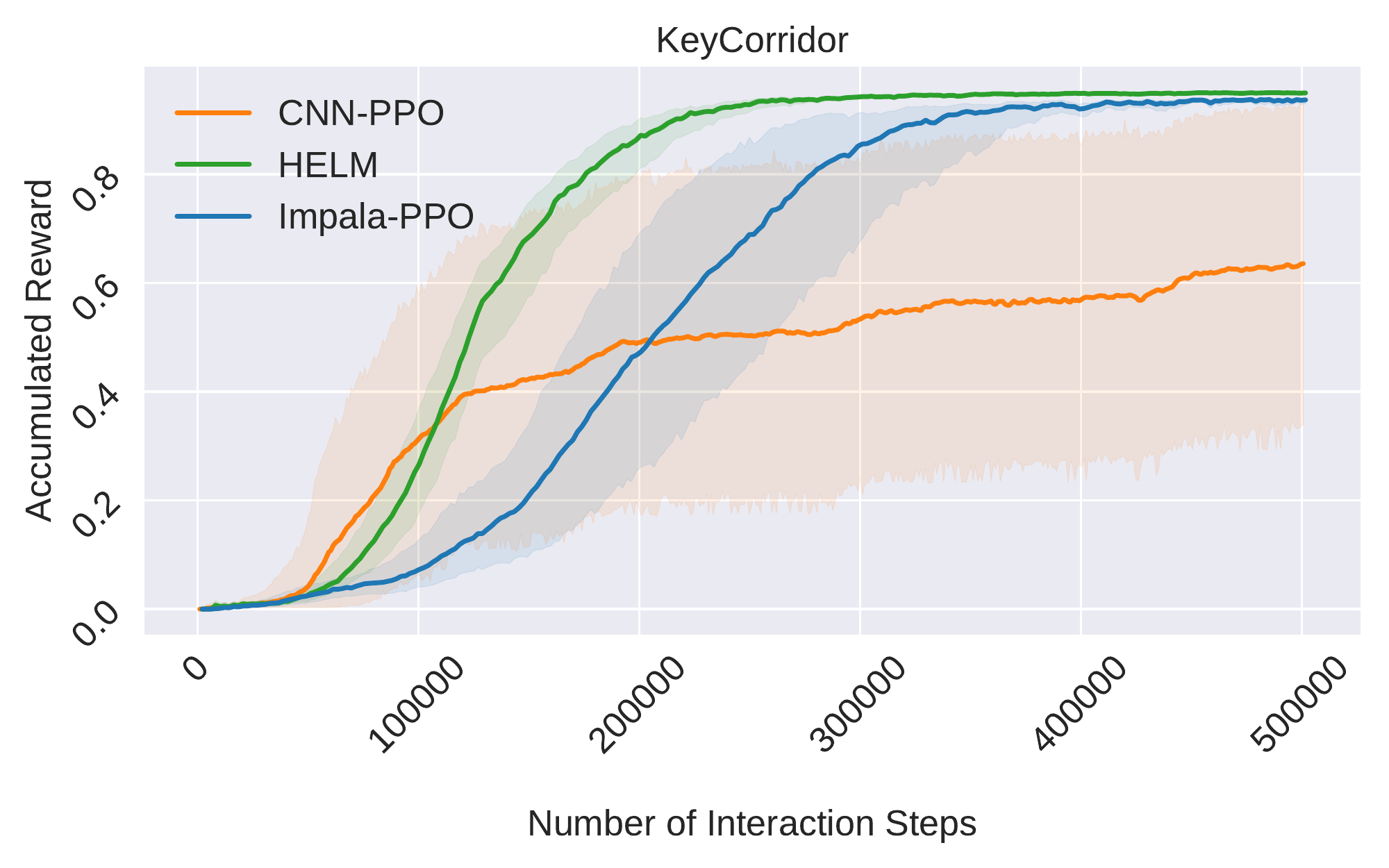}}
\caption{IQM and 95\,\% bootstrapped CIs over 100 episodes for \ourmethod, \baseline, and CNN-PPO on the \keycorr\ environment. Performance across 30 seeds is shown.}\label{fig:toytask_ppo}
\end{center}
\vskip -0.2in
\end{figure}

We limit our budget of interaction steps with the environment to 500K and show performance across 30 different seeds.  We observe that \ourmethod~significantly outperforms \baseline\ ($p=2.4\text{e-}7$; see \cref{fig:toytask_ppo}). This indicates that it successfully compresses the picked up key in its memory representation, which allows solving the environment within fewer interaction steps. 
Surprisingly, CNN-PPO is also capable of solving the environment in single runs, however exhibits large variance across multiple runs.
We provide results on 5 more partially observable Minigrid environments in the \cref{app:minigrid} and find that most of these environments are actually solvable with a Markovian policy.
Still, \ourmethod\ significantly outperforms CNN-PPO ($p=0.035$) and \baseline\ (see \cref{app:minigrid}).
This highlights the capabilities of \ourmethod~to, provide rich history representations that result in enhanced sample efficiency and improved performance even in environments where no memory component is required.

\subsection{Procgen}
Procgen provides a mode for evaluating the memory mechanism of an agent as well as sample efficiency. Out of a total of 16 environments, six support the \textit{memory} distribution mode, namely caveflyer, dodgeball, heist, jumper, maze, and miner.
The problems posed in the \textit{memory} mode are particularly challenging to solve since either observations are cropped and agent centered, or the level generation logic is altered. While Procgen provides dense rewards, it is critical that history representations are invariant to spurious correlations introduced by human-designed backgrounds.
The maximum score for each Procgen environment in the \textit{memory} mode can be found in \cref{app:procgen}. 

\begin{figure*}[ht]
\vskip 0.2in
\begin{center}
\centerline{\includegraphics[width=\textwidth]{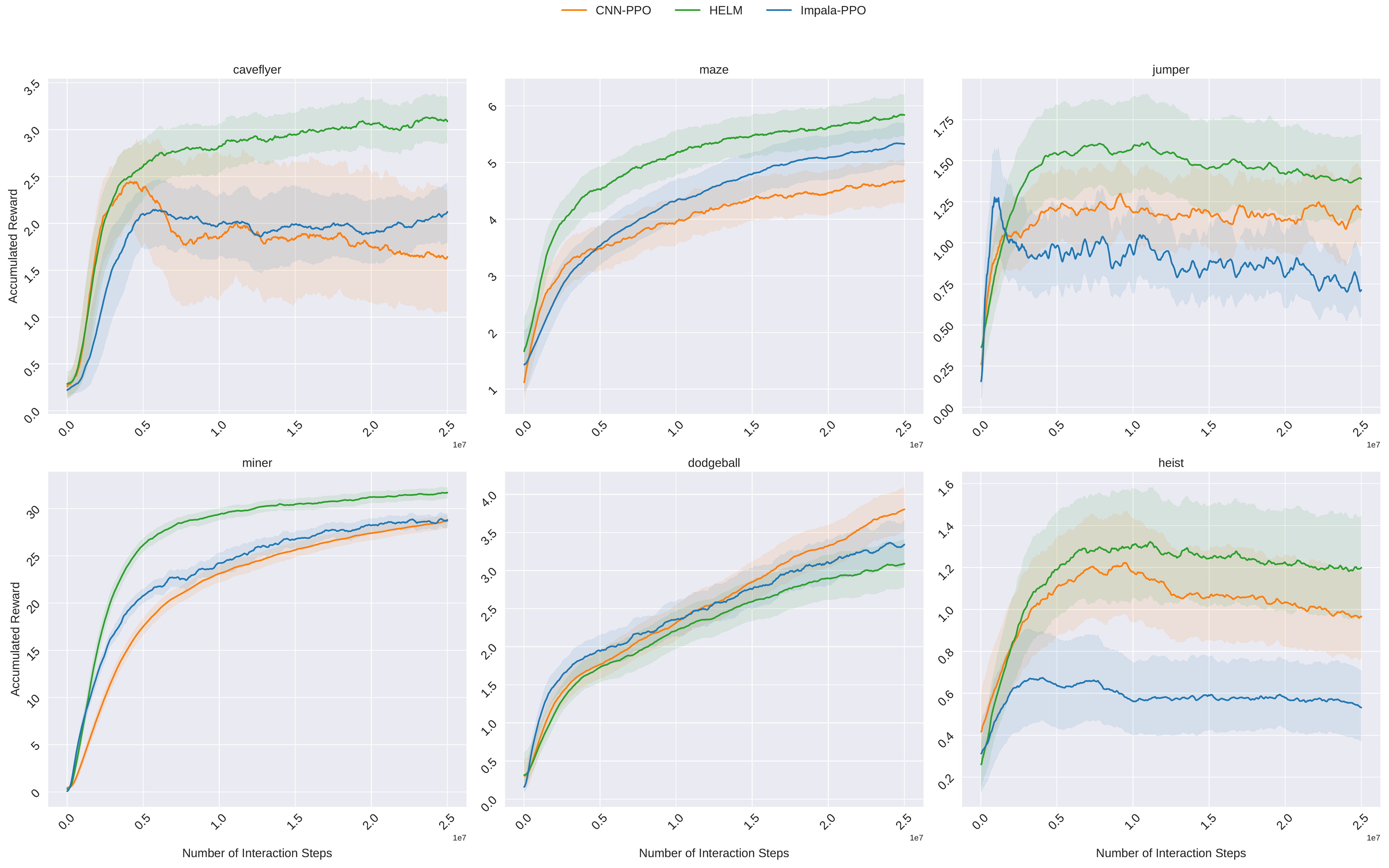}}
\caption{IQM and 95\,\% confidence intervals of the return over 100 episodes across 10 seeds for \ourmethod~and \baseline~on Procgen environments caveflyer, dodgeball, jumper, maze, miner, and heist for distribution mode \textit{memory}.}\label{fig:procgen_results}
\end{center}
\vskip -0.2in
\end{figure*}

The budget of interaction steps for Procgen is limited to 25M steps and we train on the entire level distribution across 10 seeds to evaluate for sample efficiency.
The resulting learning curves are shown in \cref{fig:procgen_results}.
\ourmethod\ significantly outperforms \baseline\ in 4 out of 6 environments, namely miner ($p=2\text{e-}4$), caveflyer ($p=0.03$), jumper ($p=0.001$), and heist ($p=2.4\text{e-}4$). Although not significant ($p=0.052$), \ourmethod~still shows superior sample efficiency on maze.
We can confirm the findings of \citet{cobbe_leveraging_2020} that training an LSTM on the Procgen environments can lead to instabilities in certain environments (heist for \textit{hard} mode).
This effect is exacerbated in the \textit{memory} mode and affects more environments (jumper, heist, caveflyer) since environments are larger and partially observable.
To compare \ourmethod\ to a more stable baseline we add results for the CNN-PPO agent.
While CNN-PPO is more stable in jumper it still exhibits instabilities in heist and caveflyer.
\ourmethod\ however performs more stable in these environments.
Remarkably, the CNN-PPO agent significantly outperforms both \ourmethod\ ($p=6.5\text{e-}4$) and \baseline\ ($p=0.003$) in the dodgeball environment.
This is due to the fact that none of the agents reached the stage of the game in which memory is actually required (finding the way back to the goal tile) and only need to perform reactive tasks (shooting enemies that spawn next to it and dodge incoming shots).
In most other environments the addition of a memory component results in significant improvements of HELM over CNN-PPO ($p=5.4\text{e-}3$ for heist, $p=0.016$ for caveflyer, $p=3.8\text{e-}4$ for miner, $p=6.2\text{e-}4$ for maze).

\begin{figure}[ht]
\vskip 0.2in
\begin{center}
\centerline{\includegraphics[width=\columnwidth]{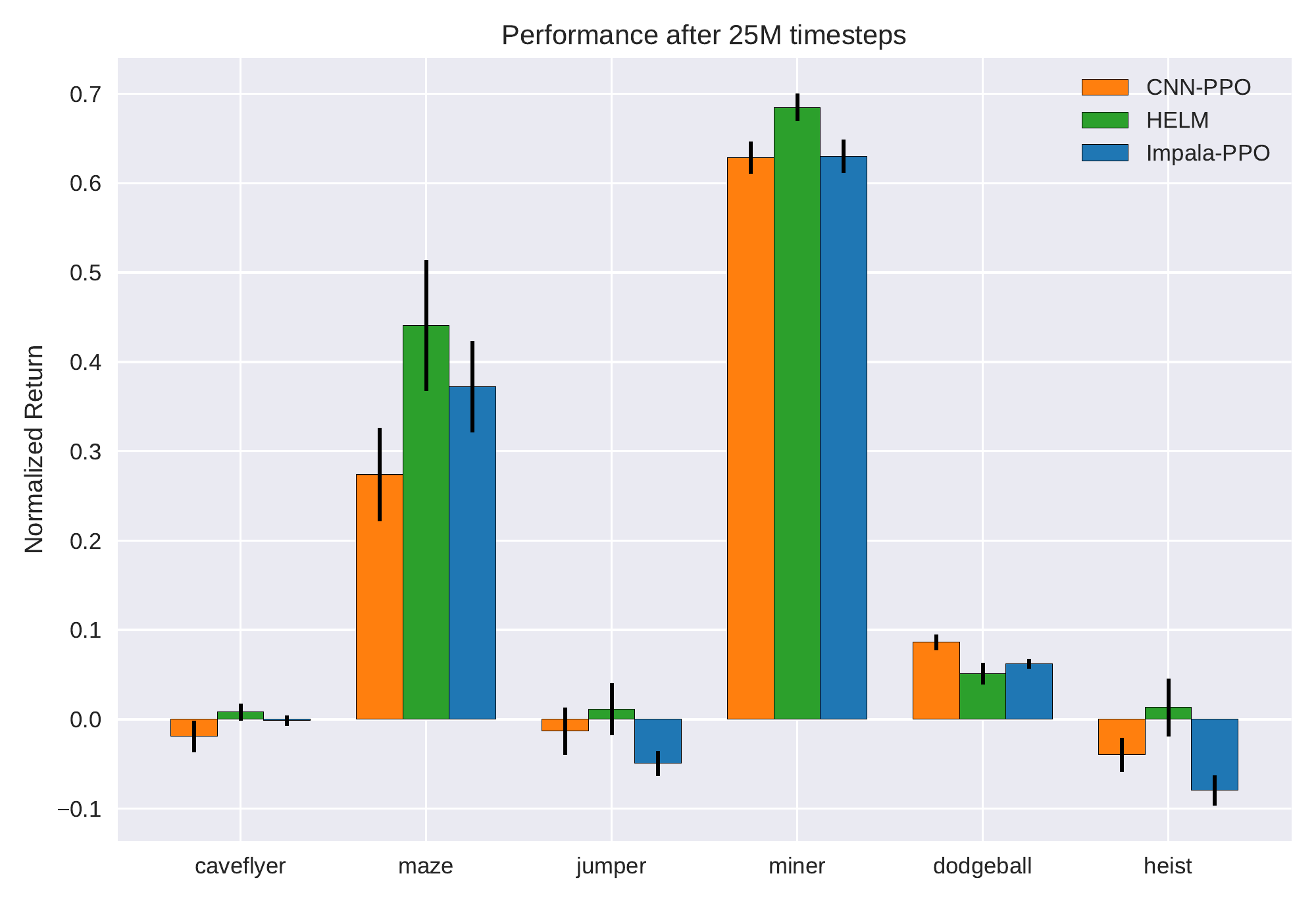}}
\caption{IQM and 95\,\% bootstrapped CIs of the normalized return across 10 seeds over the last 100 episodes for all Procgen environments. Returns are normalized according to ranges in \cref{app:procgen}.}
    \label{fig:normal_return_final}
\end{center}
\vskip -0.2in
\end{figure}

\cref{fig:normal_return_final} shows the normalized return for each environment at the end of training. 
The upper bound for the normalization range is determined by the maximum possible score for each environment. 
The lower bound is determined by training \baseline~on fully masked observations to determine what score is trivially reachable. 
Remarkably on jumper and heist, both \baseline\ and CNN-PPO even exhibit a negative normalized return.
Additionally, CNN-PPO reaches a negative normalized return on caveflyer.
A possible reason for that might be the presence of spurious correlations in the backgrounds and the need for longer training to deal with them accordingly.
When observations are masked out the agent solely relies on positional information and is not affected by noise in the observations  which leads to improved performance over CNN-PPO and \baseline\ within the 25M interaction steps.
Nevertheless, \ourmethod~manages to progress in heist and reaches a positive normalized return on jumper which highlights the stability of \ourmethod~over the baselines on particularly hard environments.

\begin{figure}[ht]
\vskip 0.2in
\begin{center}
\centerline{\includegraphics[width=.9\columnwidth]{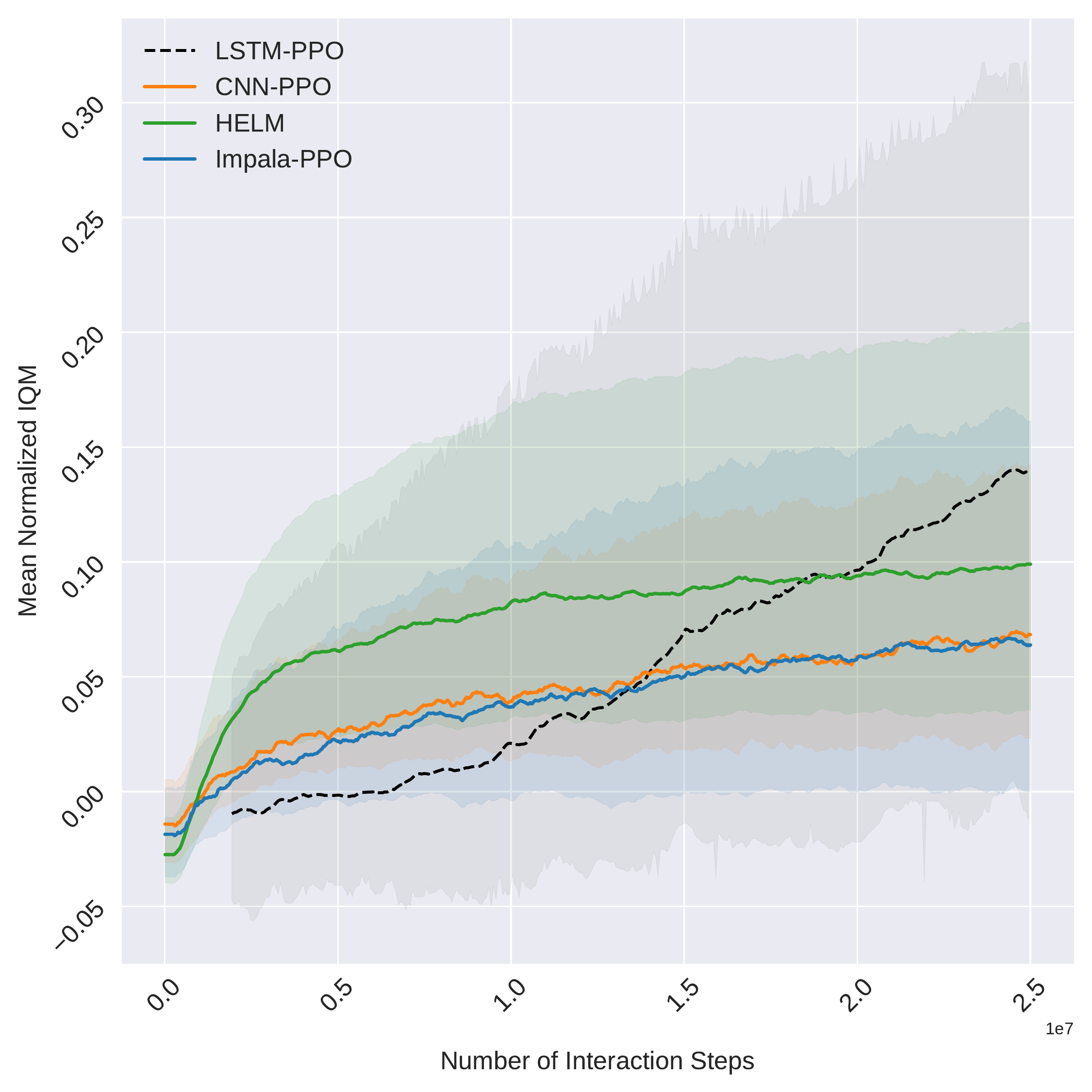}}
\caption{Mean IQM and 95\,\% bootstrapped CIs of normalized return across the six Procgen environments over 10 seeds. LSTM-PPO refers to the benchmark results from \citet{cobbe_leveraging_2020} on distribution mode \textit{hard}, which provides fully observable states. \ourmethod~significantly outperforms \baseline.}
    \label{fig:meannormscore}
\end{center}
\vskip -0.2in
\end{figure}

We plot the mean IQM (see \cref{fig:meannormscore}) across all 6 Procgen environments.
Although both methods show high variance due to low scores in particularly hard environments, like jumper, \ourmethod\ significantly outperforms \baseline\ ($p=0.022$) while showing improved sample efficiency compared to CNN-PPO ($p=0.058$).
We include a comparison to the LSTM baseline trained by \citet{cobbe_leveraging_2020} on distribution mode \textit{hard} (LSTM-PPO). 
In the \textit{hard} mode the state of the environment is fully observable, thus, one would suspect the performance of a recurrent agent trained on the \textit{hard} mode to significantly exceed the performance of an agent trained on the \textit{memory} mode.
Surprisingly, we observed that LSTM-PPO performs on-par with \ourmethod\ ($p=0.61$).
Furthermore, \ourmethod~achieves the highest mean normalized IQM across all Procgen environments on the \textit{memory} mode.

\subsection{Ablation Studies}\label{sec:ablations}

We compare \ourmethod~to:
\begin{itemize}
    \item Training a linear projection as input to the Transformer without \textit{\frozenpool}\ (HELM-TrainInput)
    \item Training the linear projection within \textit{\frozenpool} (HELM-TrainFH)
    \item The Frozen Pretrained Transformer (FPT) setting analogous to \citet{lu_pretrained_2021} by training an input projection, and finetuning layer norm parameters \citep{ba_layer_2016}, and the actor-critic networks (HELM-FPT)
    \item Training an input projection and fully finetuning TrXL (HELM-Finetuned)
\end{itemize}

\begin{figure}[ht]
\vskip 0.2in
\begin{center}
\centerline{\includegraphics[width=\columnwidth]{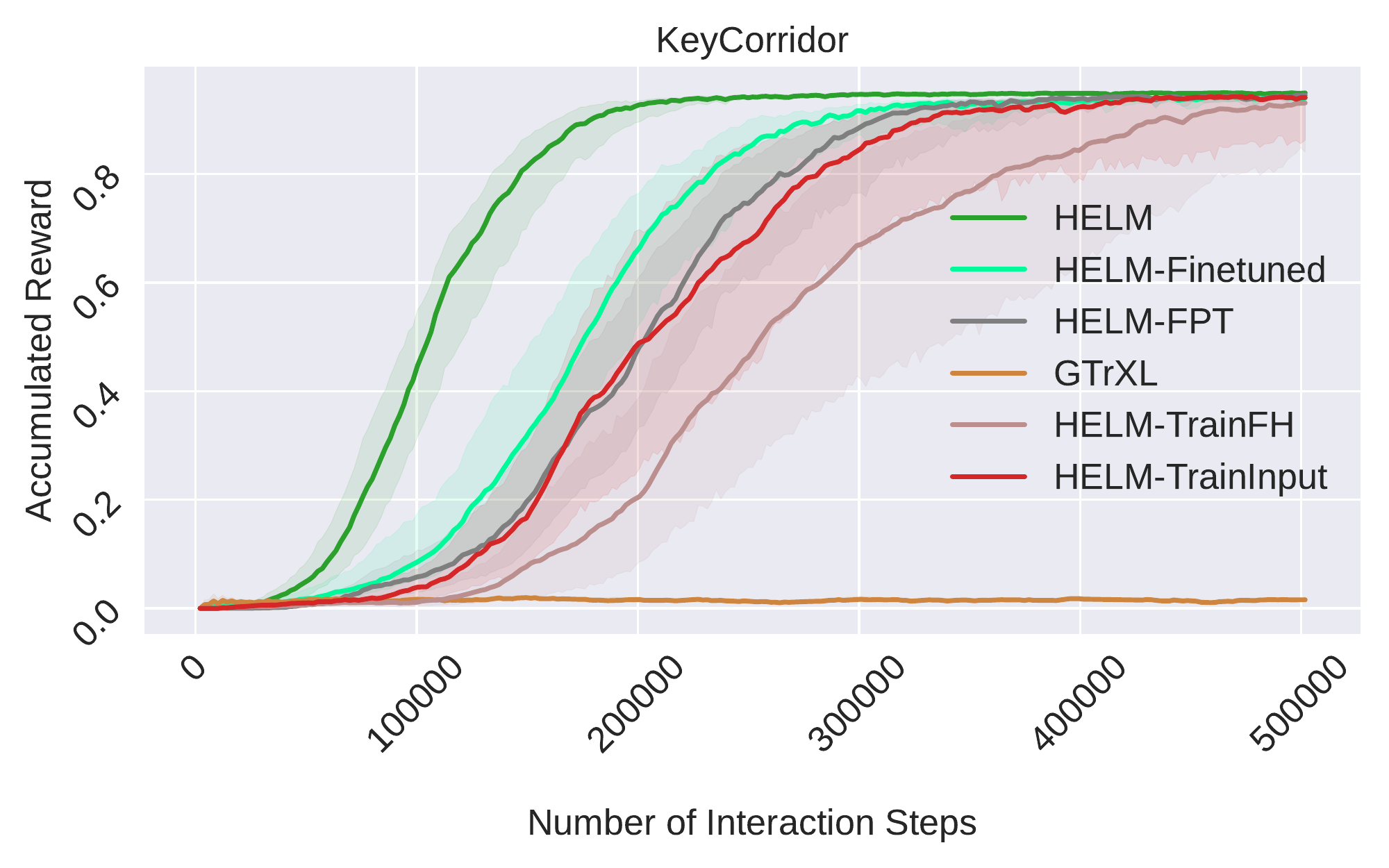}}
\caption{Comparison of \ourmethod~to various finetuning techniques and a GTrXL trained from scratch. \ourmethod~significantly outperforms all competitors. IQM and 95\,\% bootstrapped confidence interval of the average return over 100 episodes across 30 seeds are shown.}\label{fig:main_ablation}
\end{center}
\vskip -0.2in
\end{figure}

Since \citet{rothermel_dont_2021} found the learning rate to be critical for finetuning PLTs for cross-modal transfer, we run all aforementioned experiments with three different values for the learning rate used to finetune the memory mechanism (see \cref{app:ablation}). 
The learning rate for the remaining trainable parameters is set to the best value found by the hyperparameter search for \ourmethod\ (see \cref{app:hyperparam}). 
Furthermore, we include a comparison to the gated TransformerXL (GTrXL, \citealp{parisotto_efficient_2021}). 
For a fair comparison we initialize GTrXL to approximately match the number of parameters and the hidden dimensionality of TrXL and perform a hyperparameter search (\cref{app:hyperparam}).
The best learning rate for each setting is chosen and compared to \ourmethod\ on the \keycorr\ environment (\cref{fig:main_ablation}).
Additionally, we show the performance of all aforementioned variants on the Procgen environment maze for different learning rates (see \cref{app:ablation}).

Remarkably, \ourmethod\ significantly ($p=5.1\text{e-}4$) outperforms all aforementioned methods in terms of sample efficiency after training for 500K timesteps. 
GTrXL fails to make any significant progress in the environment. 
This might be explained by the difficulty of training or finetuning in the low-data regime of on-policy RL since a distribution shift occurs every time the policy is updated.
Previous work \citep{zhang_revisiting_2021,mosbach_stability_2021} highlighted problems with finetuning PLTs in low-data regimes and proposed longer training times as remedy. 
The same applies to GTrXL, since our budget of interaction steps with the environment is much smaller than the one in \citet{parisotto_efficient_2021}.
Furthermore, we observe that fully finetuning the PLT (HELM-Finetuned) is more sample efficient than re-training an input projection (HELM-TrainInput), or minimal finetuning (HELM-FPT), which is in line with findings of \citep{rothermel_dont_2021}.
However, we observe that a frozen PLT already provides compact history representations which enables solving the \keycorr~environment with fewer interaction steps. 

We conduct additional ablation studies on exchanging TrXL with LSTM models pretrained on language modeling and dynamics modeling in the \keycorr~environment.
We find that \ourmethod\ even yields performance on-par with an agent that uses a dynamics model pretrained in the \keycorr\ environment.
Further we provide compelling evidence on the importance of transferring from the language domain.
The results for our ablation studies can be found in \cref{app:ablation}.

\subsection{Analysis of representations}
\label{sec:analysis}
We conduct a qualitative analysis on the abstraction obtained by the \textit{FrozenHopfield mechanism} and the history representation obtained by the PLT.
First, we verify the distance preserving property of the \textit{FrozenHopfield} component by comparing the distances of observations across different environments before and after being processed.
In this regard we sample observations with a random policy from two Minigrid environments, namely \keycorr\ and \dynobst\ and measure their distances in observation space and in the token embedding space at the output of the \textit{FrozenHopfield} component for different values of $\beta$ (\cref{fig:dist_mats}).
We find that the parameter $\beta$ is crucial to maintain the distance preserving property since it controls the amount of dispersion of the embeddings in the token space.
Even when distances are not entirely preserved after the linear projection, they can still be enhanced by setting $\beta$ to larger values. 

\begin{figure}[ht]
\vskip 0.1in
\begin{center}
\centerline{\includegraphics[width=\columnwidth]{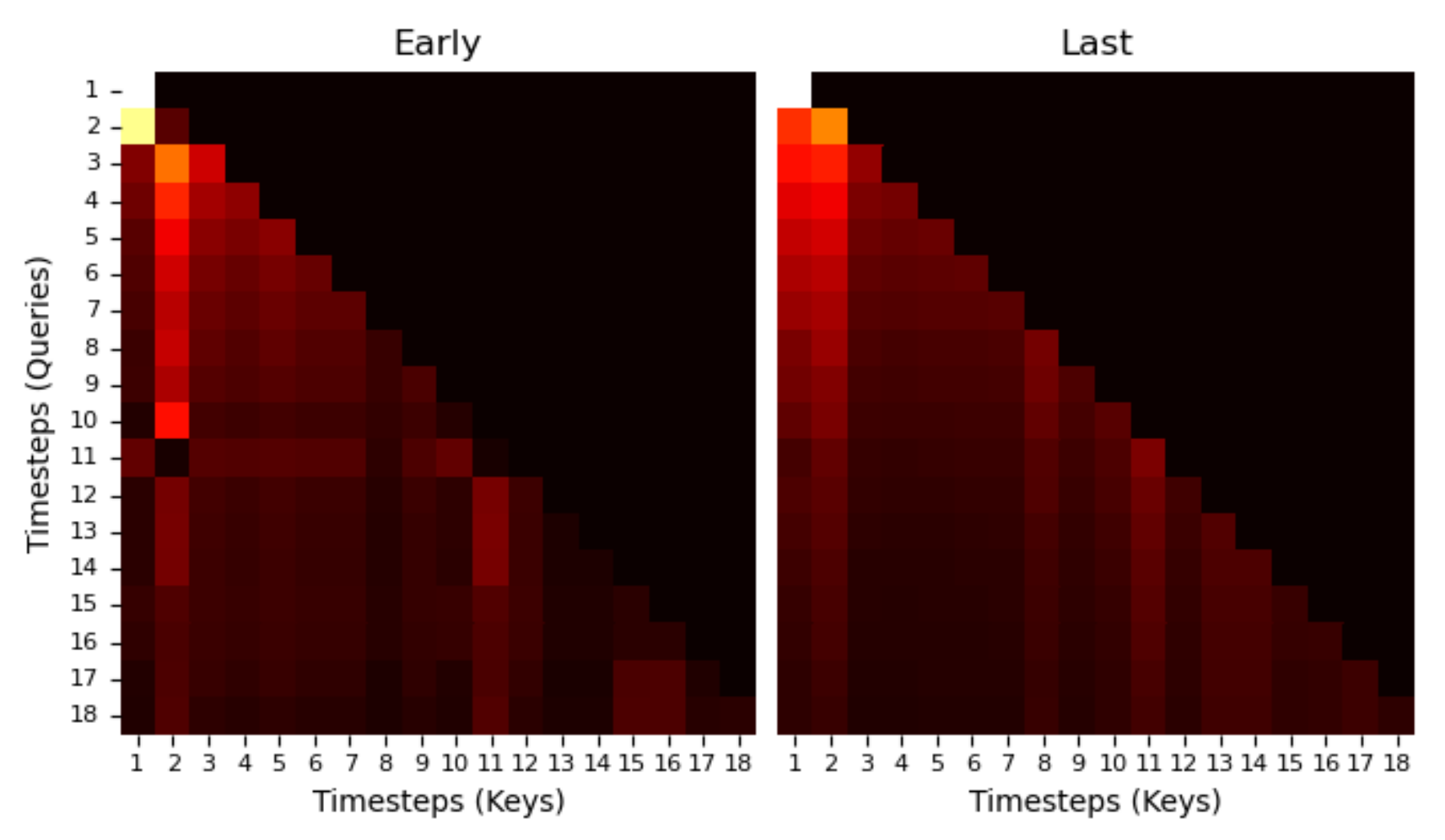}}
\caption{Attention maps of an attention head in an early and one of the last layers of TrXL for an episode sampled with a trained policy. The language model attends to changes in the observations that correspond to important events, e.g., observing the key for the first time (Timestep 2), or observing the ball for the first time (Timestep 11). }\label{fig:attn_maps}
\end{center}
\vskip -0.2in
\end{figure}

Furthermore, the attention maps of the PLT show that in certain attention heads the Transformer model attends to observations that correspond to important events for solving the task at hand.
\cref{fig:attn_maps} shows the attention maps of TrXL for an episode sampled with a trained policy consisting of 18 interaction steps with the environment.
The full episode can be observed in \cref{fig:sample_episode}.
We observe that the PLT reacts to certain changes in its input which correspond to key events in the \keycorr\ environment.
For example, at timestep 2 the agent observes the key for the first time, or at timestep 11 the agent observes the ball for the first time.
Moreover, when considering the token annotations in \cref{fig:sample_episode}, which corresponds to the closest token (i.e. $\beta \rightarrow \infty$), we find that observations that share similarity in the input space map to the same tokens, e.g., ''recollection`` when the agent faces a closed door.
Thus, the parameter $\beta$ controls the amount of clustering in the token embedding space to form abstractions of previously encountered events.

\section{Related Work}\label{sec:relwork}

Credit assignment is one of the central problems of RL \citep{puterman_markov_1994, sutton_reinforcement_2018}. Assigning credit is difficult in the presence of delayed rewards \citep{sutton_temporal_1984} and partial observability \citep{cassandra_acting_1994}. History compression has been used to address credit assignment \citep{arjona-medina_rudder_2019, patil_align-rudder_2020,holzleitner_convergence_2020,widrich_modern_2021} as well as partial observability \citep{hausknecht_deep_2015, vinyals_grandmaster_2019}.
The question of what information to store in a stream of data has been well studied before. \citet{schmidhuber_learning_1992} ignores expected values and only stores unexpected inputs. Continual learning methods like \citet{ruvolo_ella_2013} and \citet{schwarz_progress_2018} store model components over a history of tasks. Methods addressing catastrophic forgetting \citep{kirkpatrick_overcoming_2016,zenke_continual_2017} implicitly compress knowledge from previous tasks.  

The Transformer architecture \citep{vaswani_attention_2017} has dominated the field of natural language processing over the past few years \citep{devlin_bert_2019,radford_language_2019,brown_language_2020}.
Recent work adjusted the Transformer architecture for RL \citep{parisotto_stabilizing_2020,sukhbaatar_not_2021,parisotto_efficient_2021}.
Pretraining on large-scale text corpora enables such complex models to abstract relevant knowledge for solving math-word problems \citep{griffith_solving_2019}, symbolic math problems \citep{noorbakhsh_pretrained_2021}, and even university-level math exercises \citep{drori_neural_2021}. As shown by \citet{petroni_language_2019,talmor_olmpics_2020, kassner_are_2020}, PLTs can learn abstract symbolic rules and perform reasoning. 

Recently, it was shown that PLTs even transfer between languages or across modalities. 
Frozen PLTs are commonly used to extract relevant knowledge about a downstream task ~\citep{li_prefix-tuning_2021,lester_power_2021}. \citet{tsimpoukelli_multimodal_2021} learn image prompts to extract knowledge obtained during pretraining in the context of cross-domain few-shot learning \citep{adler_cross-domain_2020}.
\citet{artetxe_cross-lingual_2020,minixhofer_wechsel_2021,orhan_compositional_2021} transfer a monolingual model to different target languages via finetuning.
\citet{lu_pretrained_2021} and \citet{rothermel_dont_2021} found that pretraining on natural language is beneficial for certain cross-domain downstream tasks.

Language provides useful abstractions for decision making.
In text-based environments, PLTs have been used for proposing admissible actions \citep{yao_keep_2020}, or high-level plans via prompting \citep{huang_language_2022}. 
Language can also be a useful abstraction for exploration in visual environments via a language oracle \citep{mu_improving_2022} or multi-modal pretrained models \citep{tam_semantic_2022}.
The latent structure of language has shown benefits for policy learning in environments represented in language \citep{li_pre-trained_2022}, or across domains \citep{andreas_learning_2018} by grounding visual concepts to language.
PLTs have also shown improvements for sequence modeling in the offline RL setting \citep{reid_can_2022}. 
Furthermore, language allows one-shot acquisition of previously unseen objects in simulated environments \citep{hill_grounded_2021}.
Also, it provides useful abstraction for high-level skills \citep{sharma_skill_2022,jiang_language_2019,jacob_multitasking_2021}.
Closely related to our work, \citet{hill_human_2020} uses a PLT to encode human instructions that are compressed with visual observations in an agent's memory.
In contrast, we solely use language to represent the memory of an agent in on-policy RL with scarce data and sparse rewards. 

\textit{\frozenpool}\ is an instance of a MHN \citep{ramsauer_hopfield_2021,furst_cloob_2021,schafl_hopular_2022} that can associate image observations (or any other input) with pretrained token embeddings \citep{mikolov_linguistic_2013,wu_googles_2016,sennrich_neural_2016}.  
MHNs have already been successfully applied to 
immune repertoire classification \citep{widrich_modern_2020}, and
chemical reaction prediction \citep{seidl_improving_2022}.
The JL lemma \citep{johnson_extensions_1984} was used in the context of masked language modeling \citep{wang_linformer_2020} to reduce the context length. 
In contrast, \textit{\frozenpool}\ uses a random projection to map observations into the Transformer embedding space.

Our work also draws connections to the reservoir computing theory \citep{jaeger_echo_2001,maass_model_2002,maass_real-time_2002}. 
A reservoir, e.g., a recurrent neural network, is usually initialized at random and held fixed while only a shallow readout layer is optimized for the task at hand. 

\section{Conclusion}
\label{sec:conclusion}

Partial observability in RL requires an agent to store past information in order to approximate the true underlying state of the environment. 
Language is inherently well suited to construct abstractions and efficiently convey information.
We propose \ourmethod, a novel framework that leverages pretrained language models to compress history information for actor-critic network architectures.
To translate observations to the language domain we introduced \textit{\frozenpool}, a mechanism that acts as an associative memory and transforms a given observation to input tokens for the language model.
On memory-dependent toytasks \ourmethod~outperforms an LSTM agent trained from scratch. Further, we observe consistent improvements when applying \ourmethod~to more complex environments of the Procgen benchmark suite for which we set a new state of the art. Remarkably, \ourmethod\ outperforms training a complex Transformer model from scratch, as well as finetuning different components of the memory mechanism. 
\ourmethod\ also yields comparable performance to an agent that leverages a pretrained dynamics model for history compression.
Further, \ourmethod\ provides performance gains when applied to fully observable MDPs, highlighting the benefits of history compression in the language domain. 
The generality of \ourmethod~allows processing of any input modality using any language model as well as any RL algorithm for optimization.

\section*{Acknowledgments}
We would like to thank Karl Cobbe for providing results on the LSTM-PPO reference shown in \cref{fig:meannormscore}. The ELLIS Unit Linz, the LIT AI Lab, the Institute for Machine Learning, are supported by the Federal State Upper Austria. IARAI is supported by Here Technologies. We thank the projects AI-MOTION (LIT-2018-6-YOU-212), AI-SNN (LIT-2018-6-YOU-214), DeepFlood (LIT-2019-8-YOU-213), Medical Cognitive Computing Center (MC3), INCONTROL-RL (FFG-881064), PRIMAL (FFG-873979), S3AI (FFG-872172), DL for GranularFlow (FFG-871302), AIRI FG 9-N (FWF-36284, FWF-36235), ELISE (H2020-ICT-2019-3 ID: 951847). We thank Audi.JKU Deep Learning Center, TGW LOGISTICS GROUP GMBH, Silicon Austria Labs (SAL), FILL Gesellschaft mbH, Anyline GmbH, Google, ZF Friedrichshafen AG, Robert Bosch GmbH, UCB Biopharma SRL, Merck Healthcare KGaA, Verbund AG, Software Competence Center Hagenberg GmbH, T\"{U}V Austria, Frauscher Sensonic and the NVIDIA Corporation.
\newpage

\bibliography{references}
\bibliographystyle{icml2022}

\newpage
\appendix
\onecolumn
\numberwithin{equation}{section}

\section{Environments}
\subsection{RandomMaze}\label{app:randommaze}
\begin{figure}[p]
\vskip 0.2in
\begin{center}
\centerline{\includegraphics[width=.75\textwidth]{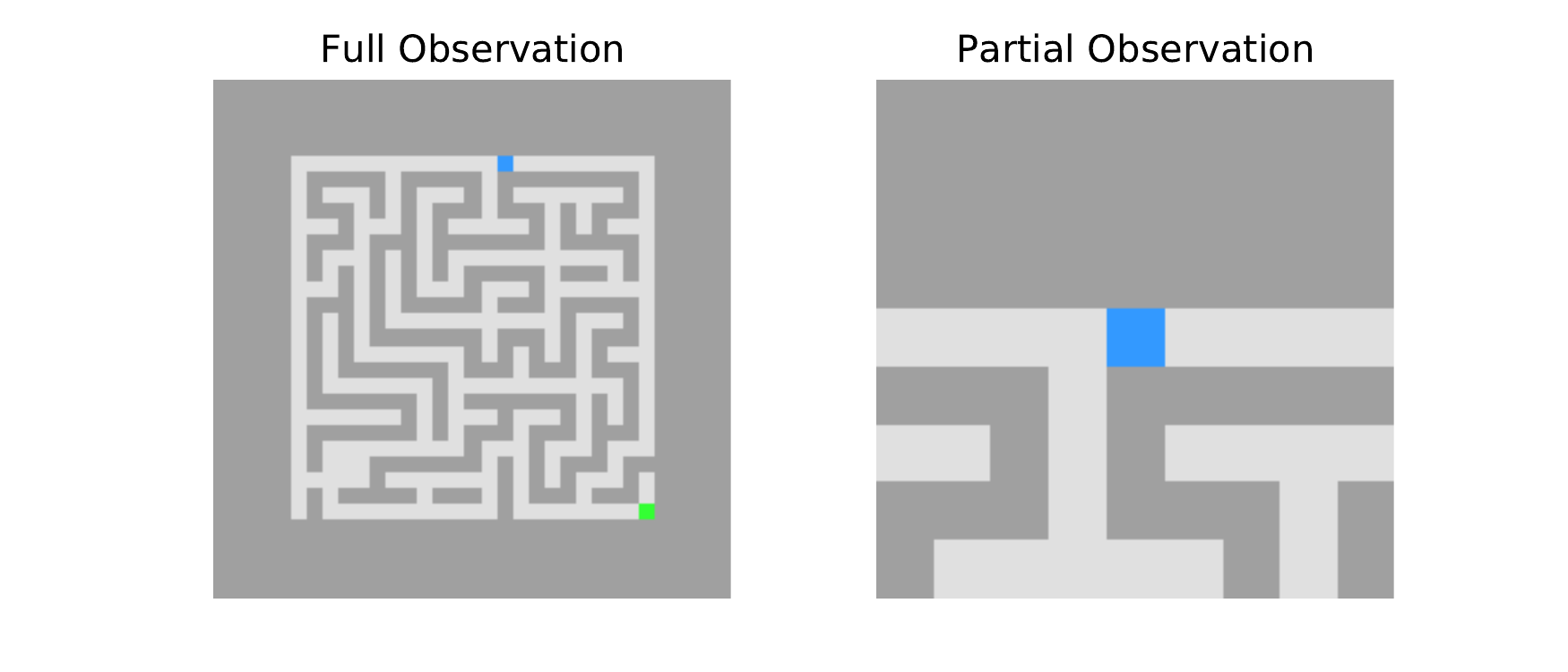}}
\caption{Example level for RandomMaze environment fully observable (left), and partially observable (right).}\label{fig:random_maze}
\end{center}
\vskip -0.2in
\end{figure}

The RandomMaze environment is implemented as a procedurally generated POMDP. Each time the environment is reset the size of the grid is sampled randomly from $\{5, \dots, 25\} \subset \dN$. The agent spawns at random positions on the grid, and the goal tile is always located in the lower right corner. Partial observability is enforced by cropping the observation to a $9 \times 9$ agent centered grid. Thus, the agent must remember if it has seen the goal before moving into different directions. Four possible actions can be issued resembling the four directions up, down, left, and right. If the agent bumps into a wall, it receives a reward of $-1$, for each interaction step it receives a reward of $-0.01$, and for reaching the goal tile it receives a reward of $+1$. \cref{fig:random_maze} depicts an example of a randomly sampled maze and its partially observable counterpart.

\subsection{Minigrid}\label{app:minigrid}
We select a set of six diverse partially observable Minigrid (\citet{chevalier-boisvert_minimalistic_2018}) environments:
\begin{itemize}
    \item \textbf{MiniGrid-\doorkey-5x5-v0 (\doorkey5x5):} The agent must pick up a key and unlock a door, before navigating to a goal tile in a $5 \times 5$ grid.
    \item \textbf{MiniGrid-\doorkey-6x6-v0 (\doorkey6x6):} Same task as above, but in a $6 \times 6$ grid, which requires more exploration.
    \item \textbf{MiniGrid-\keycorr S3R1-v0 (\keycorr):} The agent must find a key behind a closed door, unlock another door, and pick up an item behind the unlocked door.
    \item \textbf{MiniGrid-Dynamic-Obstacles-Random-5x5-v0 (\dynobst):} The agent spawns at random positions in a $5 \times 5$ grid and must navigate towards a goal tile while avoiding dynamically moving obstacles. If the agent collides with an obstacle it receives a negative reward and the episode ends. This environment provides a strong local optimum by simply avoiding the obstacles which leads to a final return of 0.
    \item \textbf{MiniGrid-Unlock-v0 (Unlock):} The agent has to open a locked door. This environment is a subtask of the \doorkey environments, but in a bigger grid.
    \item \textbf{MiniGrid-\redblue-6x6-v0 (\redblue):} The agent spawns at random positions in a $6 \times 6$ grid in which a red and a blue door is placed in opposite directions. The aim is to first open the red door and then open the blue door in that order.
\end{itemize}

\begin{figure}[p]
\vskip 0.2in
\begin{center}
\centerline{\includegraphics[width=.8\textwidth]{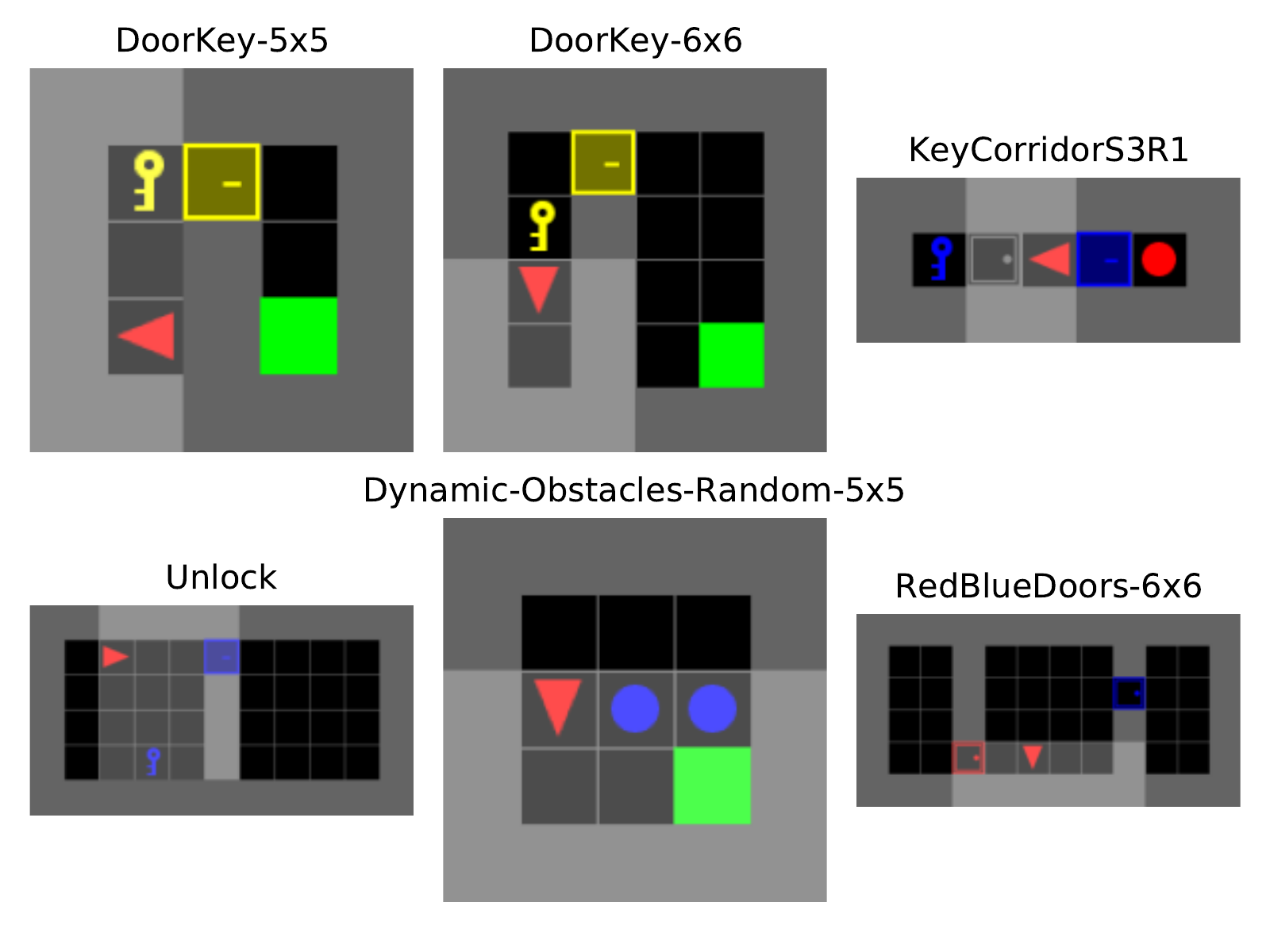}}
\caption{Example levels for Minigrid environments.}\label{fig:envplot}
\end{center}
\vskip -0.2in
\end{figure}

\begin{figure}[p]
\vskip 0.2in
\centering
\begin{subfigure}
\centering
\includegraphics[width=.55\textwidth]{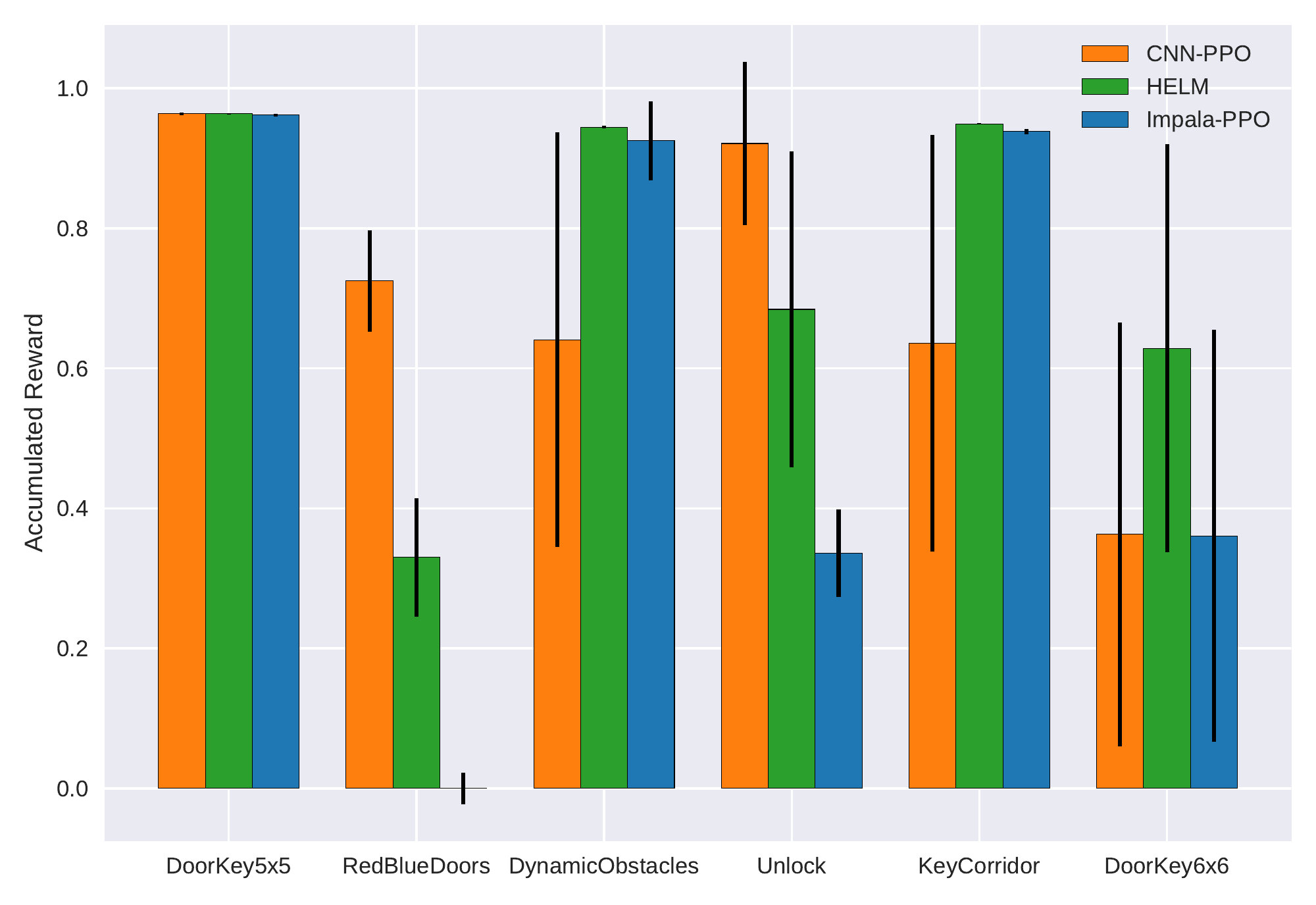}
\end{subfigure}
\hfill
\begin{subfigure}
\centering
\includegraphics[scale=.28]{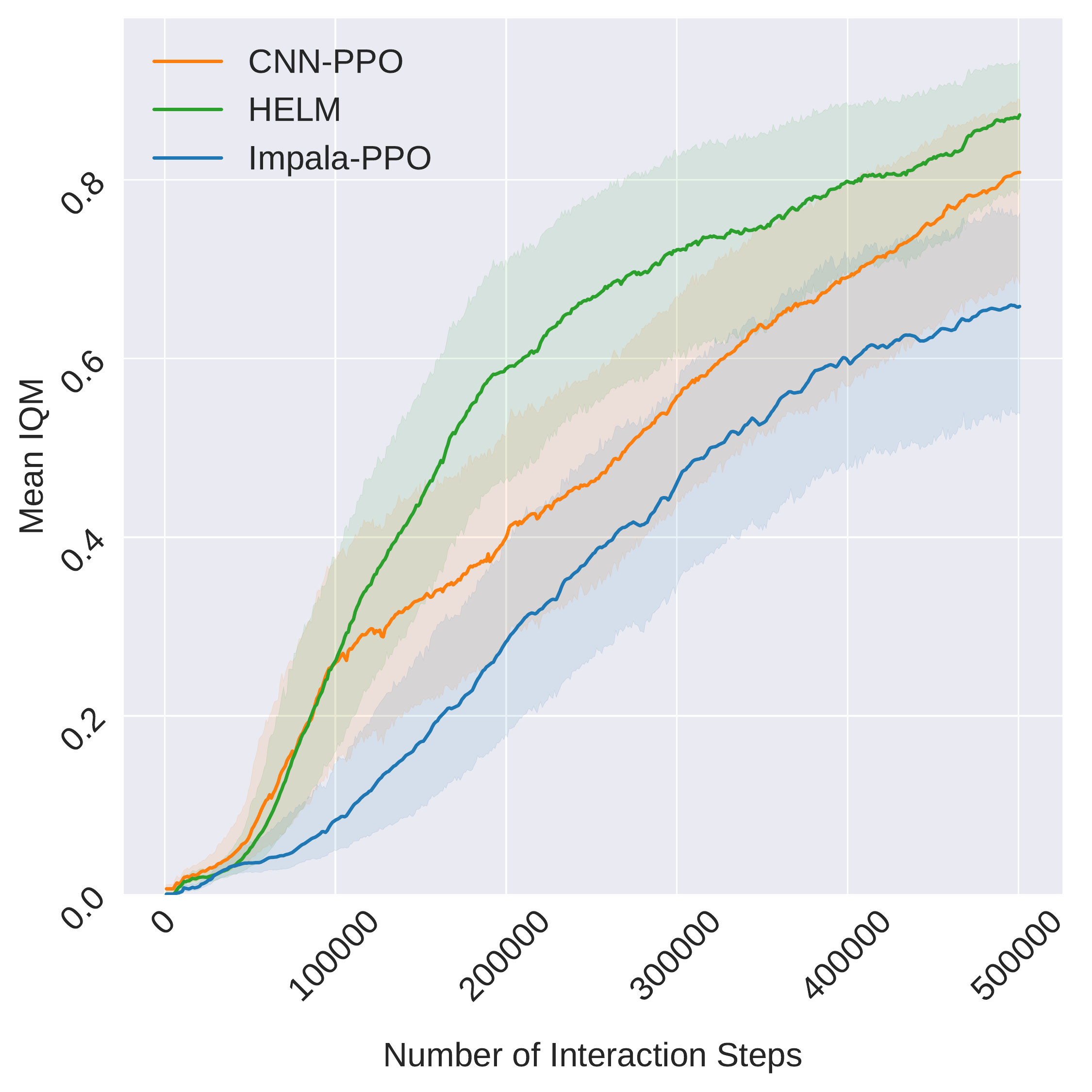}
\end{subfigure}
\caption{Comparison of return over last 100 episodes of \ourmethod\ to \baseline\ and CNN-PPO (left). IQM and 95\,\% bootstrapped CIs are shown across 30 seeds. Mean IQM and 95\,\% CIs across all environments for all methods (right).}
\label{fig:minigrid_final_all}
\vskip -0.2in
\end{figure}

\cref{fig:envplot} depicts the six different toytasks. In the partial observability setting the agent receives a $7 \times 7$ agent-centered excerpt of the grid depending on which direction the agent is currently facing. We compare \ourmethod~to a Markov policy (CNN-PPO) to elaborate whether the agent truly requires memory to solve the corresponding tasks. Further, we compare \ourmethod~to the \baseline~agent. \cref{fig:minigrid_final_all} shows the performance for each method at the end of training after 500K interaction steps. Surprisingly, we find that 5 out of 6 environments are solvable without access to memory, indicating that even though partial observability is enforced, no essential information is neglected. The only environment to be truly dependent on memory is the \keycorr~environment, for which the Markov policy stagnates at a low return. Remarkably, \ourmethod\ yields performance gains over CNN-PPO on 2 environments which are solvable without a memory mechanism while performing on par with CNN-PPO on \doorkey6x6. This highlights \ourmethod's ability of compact history compression can even be beneficial in environments where no memory is required. \baseline~is consistently outperformed by competitors.

\subsection{Procgen}\label{app:procgen}
The Procgen benchmark \citet{cobbe_leveraging_2020} allows benchmarking of RL algorithms for generalization and sample efficiency. Generalization is evaluated by training on a subset of procedurally generated environments and evaluating on a held-out set of levels. Sample efficiency is evaluated by training on the entire level distribution for a limited amount of interaction steps with the environment. Further, different distribution methods are provided to modify the training distribution, \eg \textit{easy}, \textit{hard}, or \textit{memory}. The \textit{hard} mode enhances difficulty by increasing the world size and spawning more objects as in the \textit{easy} mode. The \textit{memory} uses even larger world sizes and is only available for a subset of the environments provided by Procgen:

\begin{itemize}
    \item \textbf{caveflyer:} The agent must traverse a network of caves and reach the exit. Additional reward may be collected by shooting enemies along the way.
    \item \textbf{dodgeball:} The agent spawns in a room with a certain amount of enemies and must shoot those before unlocking a platform to which it must navigate to. 
    \item \textbf{heist:} The agent spawns in a maze and must collect a gem behind a network of locked doors. The keys for different doors are in certain colors and distributed across the environment.
    \item \textbf{jumper:} The agent must find a carrot in an open world environment which contains spike obstacles.
    \item \textbf{maze:} The agent spawns in a maze and must find a piece of cheese.
    \item \textbf{miner:} The agent must dig through dirt and collect gems before ending the episode by moving to the exit.
\end{itemize}

\begin{figure}[p]
\vskip 0.2in
\begin{center}
\centerline{\includegraphics[width=.8\textwidth]{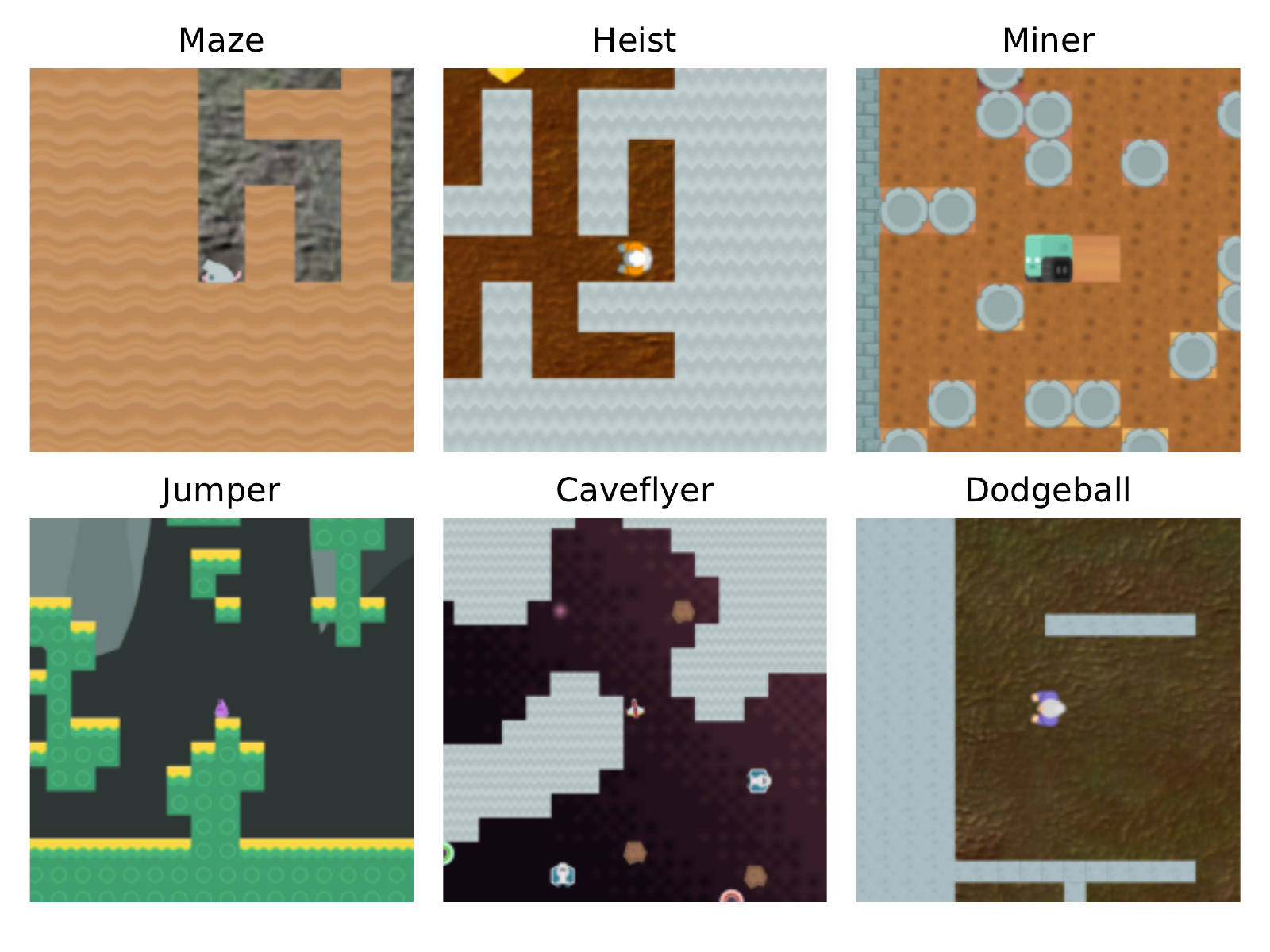}}
\caption{Procgen environments in distribution mode \textit{memory}. Observations of heist, maze, miner, and dodgeball are cropped and centered on the agent. Jumper and caveflyer avoid pruning of dead ends during game generation.}\label{fig:procgenenvs}
\end{center}
\vskip -0.2in
\end{figure}

Additonally, partial observability is enforced by cropping observations (miner, maze, dodgeball, heist), or altering the level generation logic (caveflyer, jumper). We train on the \textit{memory} distribution mode since the agent must exploit representations of the pretrained language models to effectively solve the environments. A sample observation for each of the 6 environments is shown in \cref{fig:procgenenvs}.  

The lower bound for the computation of the mean normalized return measure $R_{min}$ is obtained by training the \baseline\ agent on fully masked observations across 10 seeds and averaging over performance at the end of training. For miner, we compute the optimal upper bound $R_{max}$. For maze, jumper, and heist the $R_{max}$ value is fixed, and for caveflyer and dodgeball we estimate $R_{max}$ by sampling 100 different level configurations and averaging over their maximum reachable returns.  \cref{tab:minmaxscores} depicts all values for $R_{min}$ and $R_{max}$ for their respective environments. For LSTM-PPO, we use the originally proposed normalization ranges of the \textit{hard} mode \citep{cobbe_leveraging_2020}.

\begin{table}[ht]
\caption{$R_{min}$ and $R_{max}$ values used for computation of mean normalized score across Procgen environments for distribution mode \textit{memory}.}\label{tab:minmaxscores}
\vskip 0.15in
\begin{center}
\begin{small}
\begin{sc}
\begin{tabular}{lcc}
\toprule
 \textbf{Environment}  & $R_{min}$  & $R_{max}$\\
\midrule
caveflyer & 2.5 & 43.1 \\
dodgeball  & 1.5 & 32.1 \\
miner  & 1.3 & 46 \\
maze  & 2.7 & 10 \\
heist  & 1.2 & 10 \\
jumper  & 1.2 & 10 \\
\bottomrule
\end{tabular}
\end{sc}
\end{small}
\end{center}
\vskip -0.1in
\end{table}

\section{Hyperparameter Search}\label{app:hyperparam}
We perform a parameter search over learning rate in $\{3\text{e-}4, 1\text{e-}5, 5\text{e-}5, 1\text{e-}4\}$, entropy coefficient in $\{0.05, 0.01, 0.005, 0.001\}$, rollout length in $\{64, 128, 256\}$, and the softmax scaling factor $\beta \in \{0.5, 1, 10, 50, 100\}$ for \ourmethod~for the Minigrid environments and the RandomMaze environment. For Procgen environments, we reduce the possible values for the entropy coefficient to $[0.01, 0.005, 0.001]$ and for $\beta$ to $\{1, 10, 100\}$. \citet{cobbe_leveraging_2020} found that scaling up the number of channels of the convolutional encoder led to significant improvements. Thus, by default we scale the number of channels of the CNN encoder by $\lambda_c = 4$. For all experiments with \ourmethod, we set the number of minibatches to 8. For the \baseline~baseline we search over batch sizes of $\{4, 8, 16\}$ sequences per batch, learning rates $\{1\text{e-}5, 5\text{e-}5, 1\text{e-}4\}$, and entropy coefficients $\{0.05, 0.01, 0.005, 0.001\}$ for Minigrid and RandomMaze environments. Further, we also search over channel scaling factors $\lambda_c \in \{1, 2, 4\}$. For Procgen, we set $\lambda_c = 4$ and run a reduced search over parameter values for the entropy coefficient in $\{0.01, 0.005, 0.001\}$ and batch sizes of $\{8, 16\}$ sequences per batch. The values for $\gamma$ and $\lambda_{GAE}$ were found experimentally for Minigrid environments, while for Procgen they were chosen according to prior work \citep{cobbe_leveraging_2020}. The number of PPO epochs is set to 3 for all experiments. We found that learning rate decay did not lead to improvements except for the \redblue~environment. We also experimented with decaying the clipping range without observing any benefits though. Thus, we set the clipping range to $0.2$ for all experiments. We choose AdamW \citep{loshchilov_decoupled_2019} as optimizer for all environments with default values for weight decay. Moreover, we clip the norm of the gradients to $0.5$ for all of our experiments. For CNN-PPO, we run a grid comprising the same parameters as for \ourmethod, except for the parameter $\beta$, which is only available for the \frozenpool~component. We set $\lambda_c = 4$ by default for all environments. The number of environments for RandomMaze and Minigrid are set to 16, while it is set to 64 for Procgen. The parameter search is run for 60K, 500K, and 2.5M interaction steps for Minigrid, RandomMaze, and Procgen, respectively. We select the parameter combination of the setting which yields the highest IQM at training end over 5 seeds. The best hyperparameters for \ourmethod~for all Minigrid environments are shown in  \cref{tab:minigridhyper} while the best hyperparameters for the Procgen environments are shown in \cref{tab:procgenhyperparam}. 
For GTrXL, we perform a parameter search over the same grid as for \ourmethod, excluding the parameter $\beta$.
We use the same CNN encoder as for all other experiments for encoding the environment observations before feeding them into GTrXL.
The finetuning experiments use the same parameter values as found in the parameter search for \ourmethod, but we additionally search over different values for the learning rate $\{1\text{e-}4, 1\text{e-}5, 1\text{e-}6\}$ used for finetuning the Transformer and the input mapping. During training and finetuning we chunk sequences into a length of 20 timesteps and feed them into the Transformer after applying the input encoder. 

\begin{table}[t]
\caption{Hyperparameters for all six MiniGrid environnments, DK is short for \doorkey, and RBD is short for \redblue.}\label{tab:minigridhyper}
\vskip 0.15in
\begin{center}
\begin{small}
\begin{sc}
\begin{tabular}{l|cccccc}
\toprule
    \textbf{Parameter}  & \textbf{DK5x5}  & \textbf{DK6x6} & \textbf{Unlock}  & \textbf{\keycorr} & \textbf{\dynobst}  & \textbf{RBD}\\
\midrule
    Learning Rate & 1e-4 & 3e-4 & 3e-4 & 1e-4 & 1e-4 & 5e-5\\
    Rollout length  & 64 & 64 & 64 & 128 & 64 & 256\\
    Entropy Coeff.  & 1e-2 & 1e-2 & 1e-2 & 5e-2 & 5e-3 & 1e-2\\
    $\gamma$  & 0.99 & 0.99 & 0.99 & 0.99 & 0.99 & 0.99 \\
    $\lambda$  & 0.99 & 0.99 & 0.99 & 0.99 & 0.99 & 0.99 \\
    $\beta$  & 1 & 10 & 50 & 100 & 100 & 1\\
    $\lambda_c$ & 4 & 4 & 4 & 4 & 4 & 4\\
\bottomrule
\end{tabular}
\end{sc}
\end{small}
\end{center}
\vskip -0.1in
\end{table}

\begin{table}[t]
\caption{Hyperparameters for Procgen environnments.}\label{tab:procgenhyperparam}
\vskip 0.15in
\begin{center}
\begin{small}
\begin{sc}
\begin{tabular}{l|cccccc}
\toprule
    \textbf{Parameter}  & \textbf{caveflyer}  & \textbf{dodgeball} & \textbf{heist}  & \textbf{jumper} & \textbf{maze}  & \textbf{miner}\\
\midrule
    Learning Rate & 1e-5 & 1e-4 & 1e-5 & 5e-5 & 5e-5 & 1e-4\\
    Rollout length  & 128 & 256 & 256 & 256 & 128 & 128 \\
    Entropy Coeff. & 5e-3 & 1e-3 & 1e-2 & 5e-3 & 5e-3 & 1e-3 \\
    $\gamma$  & 0.999 & 0.999 & 0.999 & 0.999 & 0.999 & 0.999 \\
    $\lambda$  & 0.95 & 0.95 & 0.95 & 0.95 & 0.95 & 0.95 \\
    $\beta$  & 1 & 1 & 100 & 1 & 1 & 1\\
\bottomrule
\end{tabular}
\end{sc}
\end{small}
\end{center}
\vskip -0.1in
\end{table}

\begin{figure}[p]
\vskip 0.2in
\begin{center}
\centerline{\includegraphics[width=\textwidth]{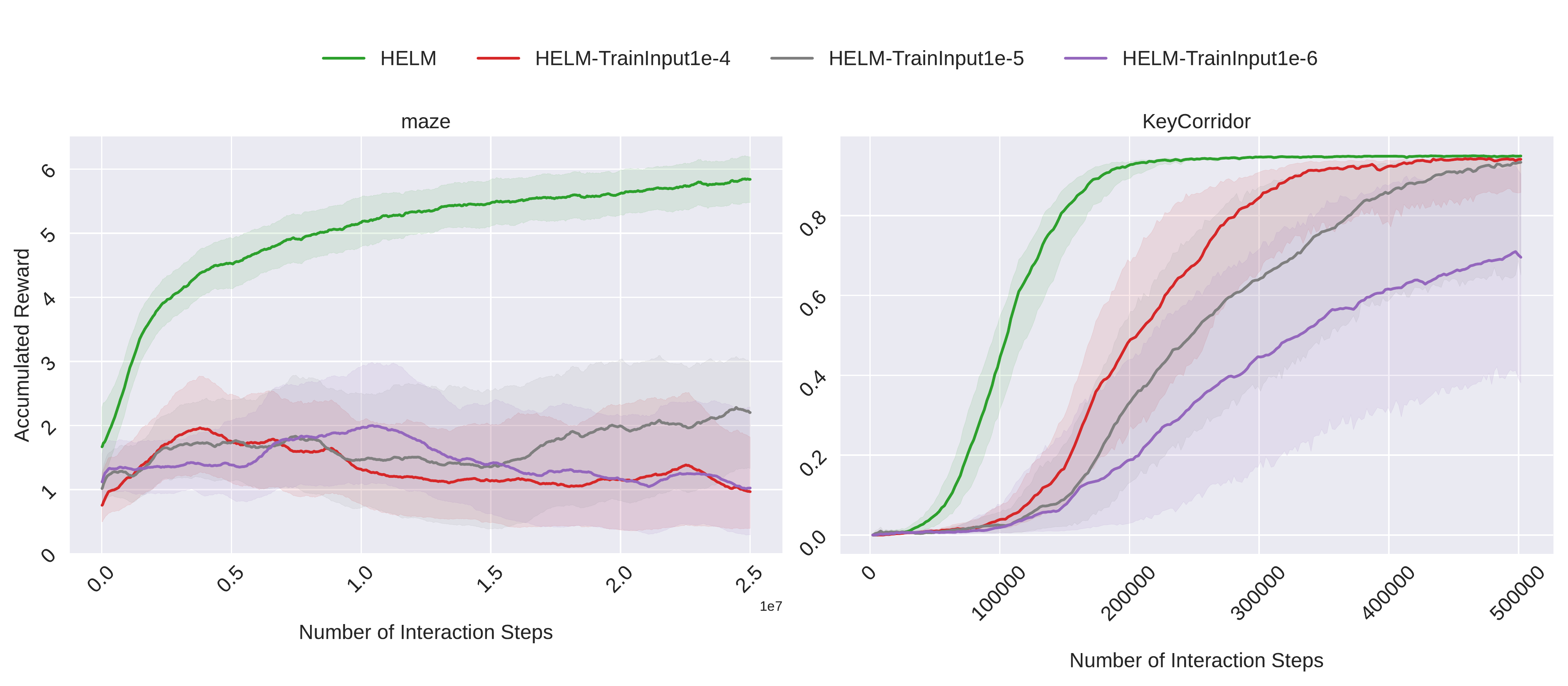}}
\caption{Comparison of \ourmethod\ to training an input mapping to the Transformer model for three different learning rates. IQM and 95\% bootstrapped confidence interval of return over last 100 episodes across 30 seeds are shown.}\label{fig:ablation_input}
\end{center}
\vskip -0.2in
\end{figure}

\begin{figure}[p]
\vskip 0.2in
\begin{center}
\centerline{\includegraphics[width=\textwidth]{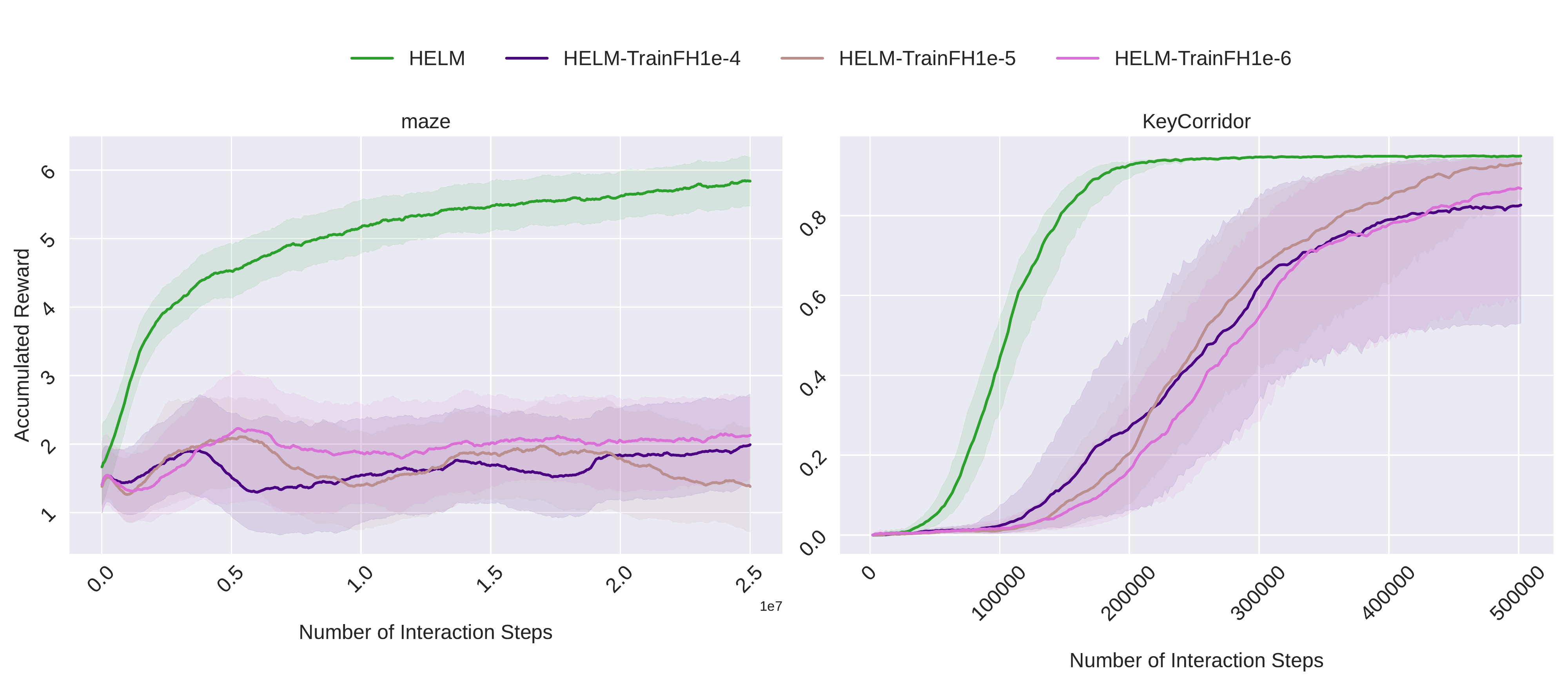}}
\caption{Comparison of \ourmethod\ to training the linear projection in the \textit{FrozenHopfield} component for three different learning rates. IQM and 95\% bootstrapped confidence interval of return over last 100 episodes across 30 seeds are shown.}\label{fig:ablation_trainfh}
\end{center}
\vskip -0.2in
\end{figure}

\begin{figure}[p]
\vskip 0.2in
\begin{center}
\centerline{\includegraphics[width=\textwidth]{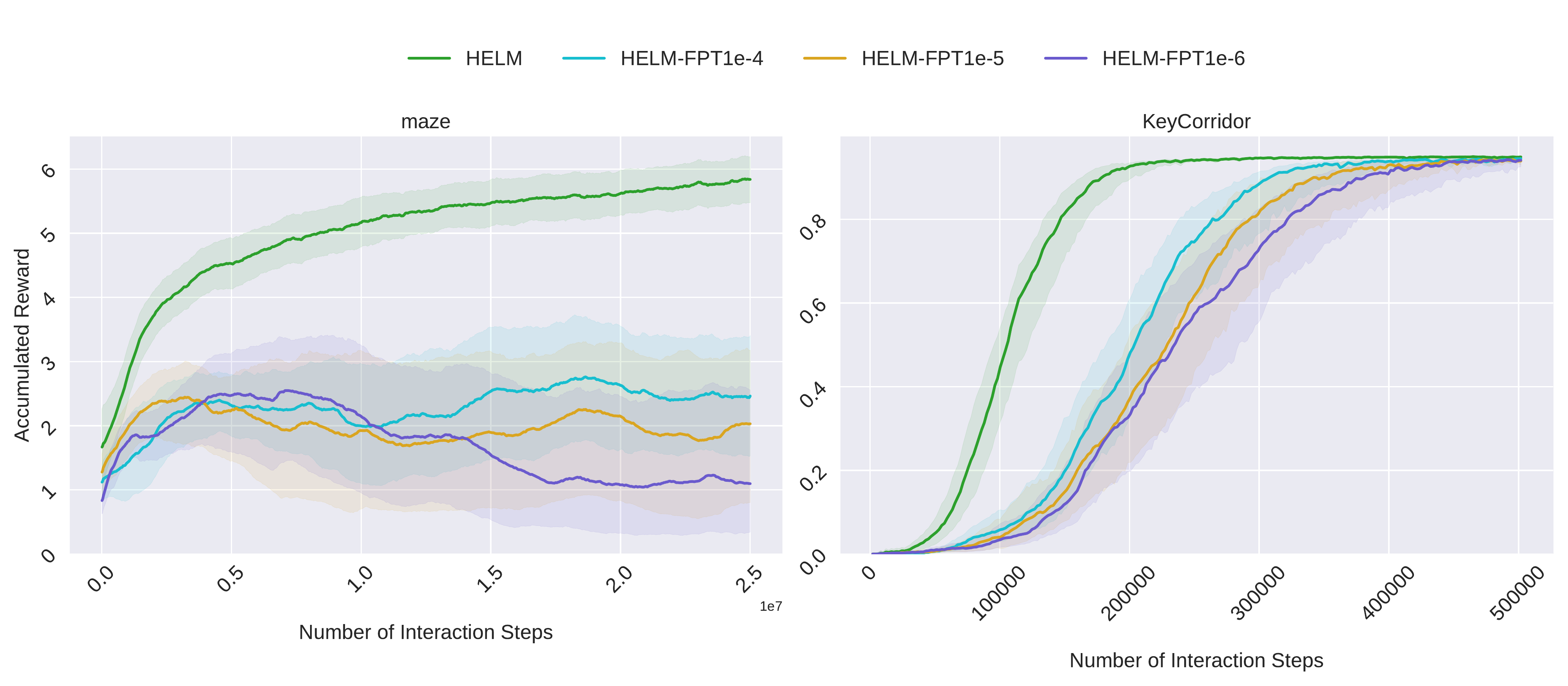}}
\caption{Comparison of \ourmethod~to the FPT setting for three different learning rates. IQM and 95\% bootstrapped confidence interval of return over last 100 episodes across 30 seeds are shown.}\label{fig:ablation_fpt}
\end{center}
\vskip -0.2in
\end{figure}

\section{PPO vs PPG}\label{app:ppo_ppg}
\begin{figure}[p]
\vskip 0.2in
\begin{center}
\centerline{\includegraphics[width=\textwidth]{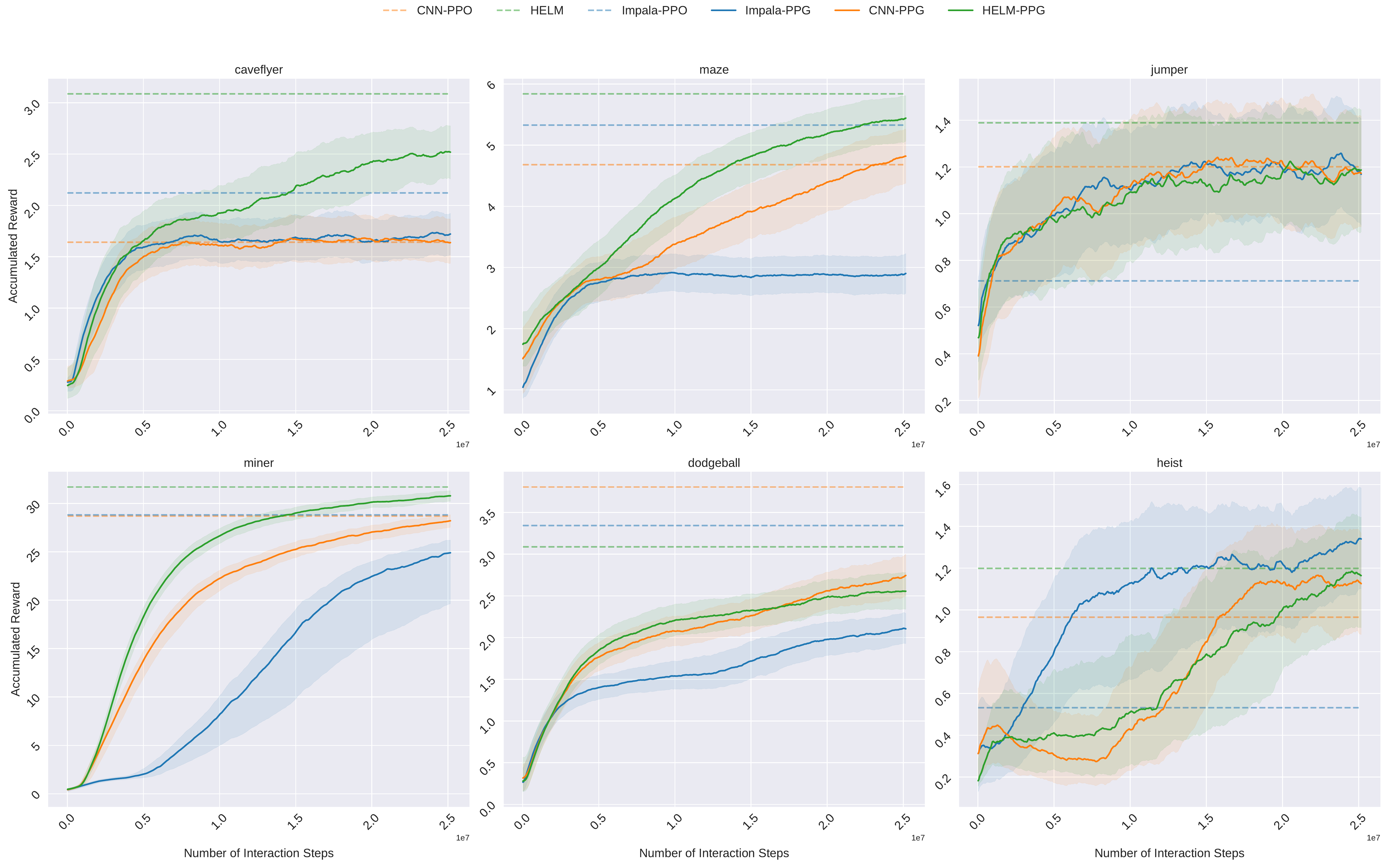}}
\caption{Performance of \ourmethod-PPG compared to Impala-PPG and CNN-PPG after training for 25M interaction steps. IQM and 95\,\% bootstrapped confidence interval of return over last 100 episodes across 30 seeds are shown.}\label{fig:procgen_ppg}
\end{center}
\vskip -0.2in
\end{figure}

\begin{figure}[p]
\vskip 0.2in
\begin{center}
\centerline{\includegraphics[width=.5\textwidth]{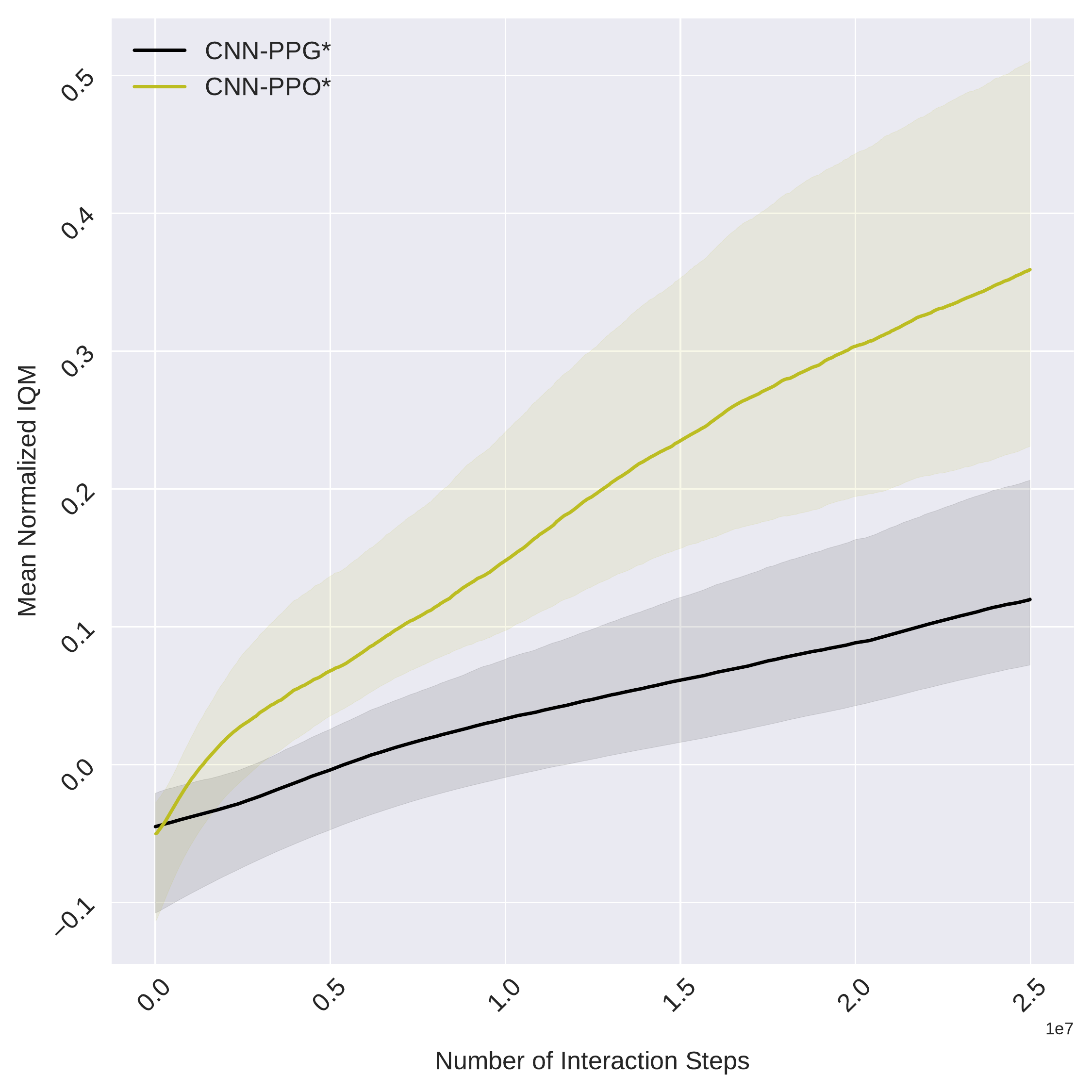}}
\caption{Mean IQM and 95\,\% bootstrapped confidence interval of the normalized return over 25M interaction steps across 3 seeds provided by the authors of PPG and PPO in their respective codebases.}\label{fig:procgen_orig}
\end{center}
\vskip -0.2in
\end{figure}

The PPG algorithm \citep{cobbe_phasic_2021} was proposed for improving sample efficiency over PPO on the Procgen benchmark suite. 
It separates the updates of the policy and the value function in different phases while adding an additional phase in which the features of the value function are distilled into the policy network.
This resulted in substantial improvements in terms of sample efficiency on the Procgen environments in the \textit{hard} mode.
Since \ourmethod\ may be optimized with any RL algorithm, we show results of our method when trained with PPG (\ourmethod-PPG).
As in \cref{sec:experiments} we compare to a recurrent agent based on the small Impala architecture (Impala-PPG) and a Markovian baseline (CNN-PPO).
Additionally, we include the performance of \ourmethod\ trained with PPO as a point of comparison.
We find that HELM-PPG significantly improves upon the baselines in 3 out of the 6 environments, namely caveflyer ($p=1.1\text{e-}3$), maze ($p=1.56\text{e-}3$), and miner ($p=6.57\text{e-}4$) (see \cref{fig:procgen_ppg}).
On the remaining environments the performance of \ourmethod-PPG\ is on-par with the competitors.
As compared to the performance using PPO we find that Impala-PPG and CNN-PPG are more stable and perform on-par with \ourmethod-PPG on heist and jumper.
However, on caveflyer and maze both Impala-PPG and CNN-PPG stagnate within the 25M interaction steps.
Furthermore, the scores reached with the PPG variants are in general lower than for PPO.
A possible reason is that PPG only overcomes PPO after sufficient amount of training, i.e., after useful features have been learned that are beneficial for the policy.
Further, it might be that the advantages of PPG are mostly observed in other Procgen environments that are not dependent on memory.
We verify this assumption by visualizing the performance of a Markovian policy trained with PPO (CNN-PPO*) and PPG (CNN-PPG*) for the 6 Procgen environments in the \textit{hard} mode across 3 seeds, which are provided by the authors in the respective codebases (\cref{fig:procgen_orig}).
Indeed, we find that the advantages of PPG over PPO can only be observed after a substantial amount of training, which exceeds the limited budget in our work.

\section{Ablation Studies}\label{app:ablation}
We perform ablation studies on the \textit{\frozenpool} mapping by exchanging it with a linear layer (\ourmethod-TrainInputX, where X is replaced by the learning rate used for training the linear mapping). As mentioned in \cref{app:hyperparam}, we conduct experiments with three different learning rates for training the encoder, while keeping the learning rate for the actor-critic head and the CNN encoder of the current timestep fixed to the learning rate found by our parameter search. 
We show results on the Minigrid \keycorr\ and Progen maze environments in \cref{fig:ablation_input}.
On both environments \ourmethod\ significantly outperforms the second-best \ourmethod-TrainInputX variant ($p=8.64\text{e-}7$ for \keycorr, $p=9.08\text{e-}5$ for maze).
On \keycorr\ the performance decreases with a lower learning rate indicating that a higher learning rate is more suited for re-learning the input projection.
Still, there is a clear performance drop compared to \ourmethod.
However, training an input projection results in severe instabilities in maze, where a lower learning rate appears to perform better.

Another experiment investigates the effect of training the linear projection in the \textit{FrozenHopfield} mechanism.
Again, we train the linear projection with three different learning rates (HELM-TrainFHX) and show the performances on maze and \keycorr\ in \cref{fig:ablation_trainfh}.
On \keycorr\ the performance is comparable for different learning rates and significantly worse than for \ourmethod\ ($p=6.19\text{e-}7$).
On maze we observe the same trend, with \ourmethod\ significantly outperforming all competitors ($p=9.13\text{e-}5$).

Next, we compare \ourmethod~to minimally finetuning parts of the Transformer language model. 
\citet{lu_pretrained_2021} introduced \textit{Frozen Pretrained Transformer} (FPT) which finetunes the input projection, the absolute positional embeddings, the layernorm parameters, and the readout for cross-modal transfer after language pretraining. 
Analogously, we train an input projection to the Transformer as a linear layer and finetune the layernorm parameters of TrXL (\ourmethod-FPTX) with three different learning rates $\{1\text{e-}4, 1\text{e-}5, 1\text{e-}6\}$ (see \cref{fig:ablation_fpt}). 
\ourmethod\ significantly outperforms all other methods on \keycorr\ ($p=5.15\text{e-}4$) and maze ($p=1.65\text{e-}4$).

\begin{figure}[p]
\vskip 0.2in
\begin{center}
\centerline{\includegraphics[width=\textwidth]{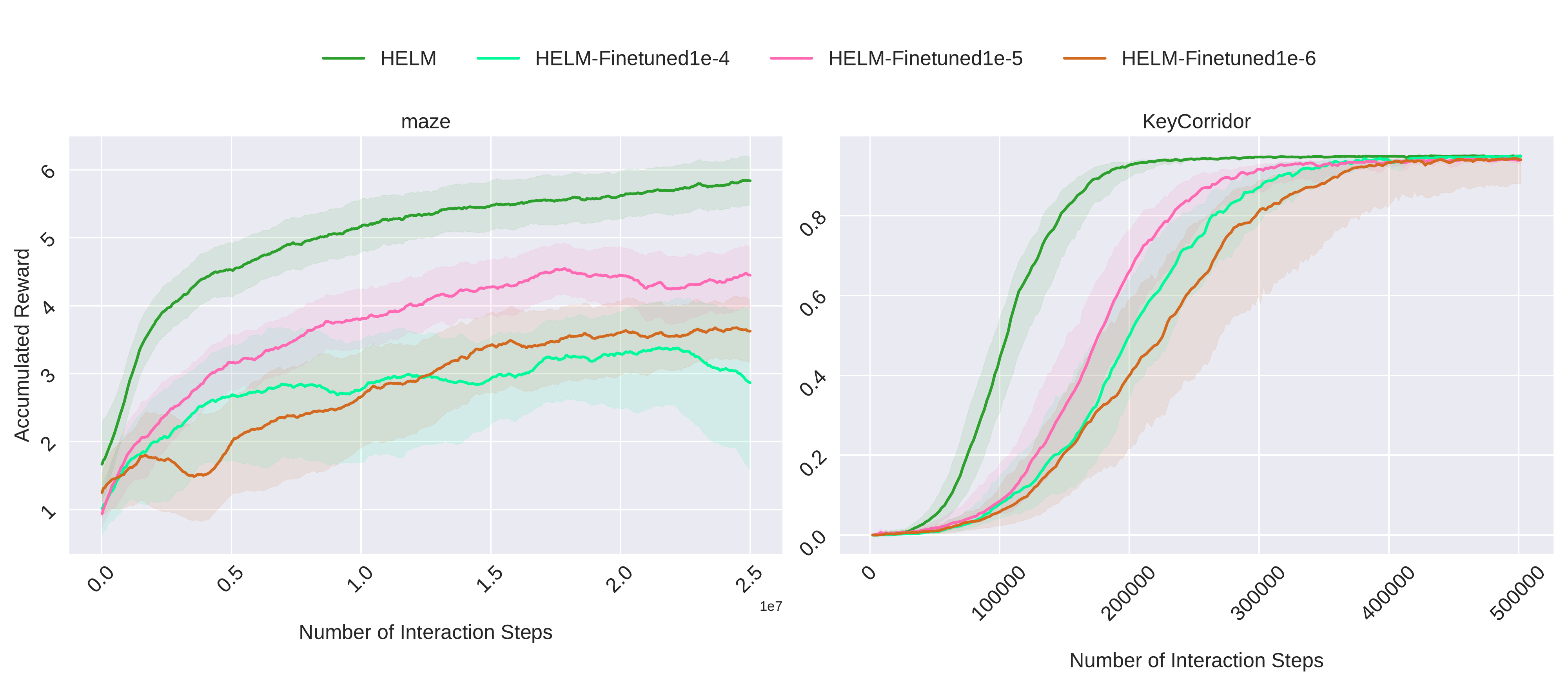}}
\caption{Comparison of \ourmethod~to fully finetuning for three different learning rates. IQM and 95\,\% bootstrapped confidence interval of return over last 100 episodes across 30 seeds are shown.}\label{fig:ablation_finetune}
\end{center}
\vskip -0.2in
\end{figure}

Moreover, we compare \ourmethod~to fully finetuning the Transformer (HELM-FinetunedX). To this end we exchange the \frozenpool~component with a linear layer and finetune all weights of the TrXL for three different learning rates ($\{1\text{e-}4, 1\text{e-}5, 1\text{e-}6\}$).
\cref{fig:ablation_finetune} shows the resulting learning curves on the \keycorr\ and the maze environments. 
\ourmethod~outperforms the second-best method on \keycorr\ and maze with a significance level of $p=7.15\text{e-}6$, and $p=1.61\text{e-}4$, respectively.

\begin{figure}[p]
\vskip 0.2in
\begin{center}
\centerline{\includegraphics[width=.75\textwidth]{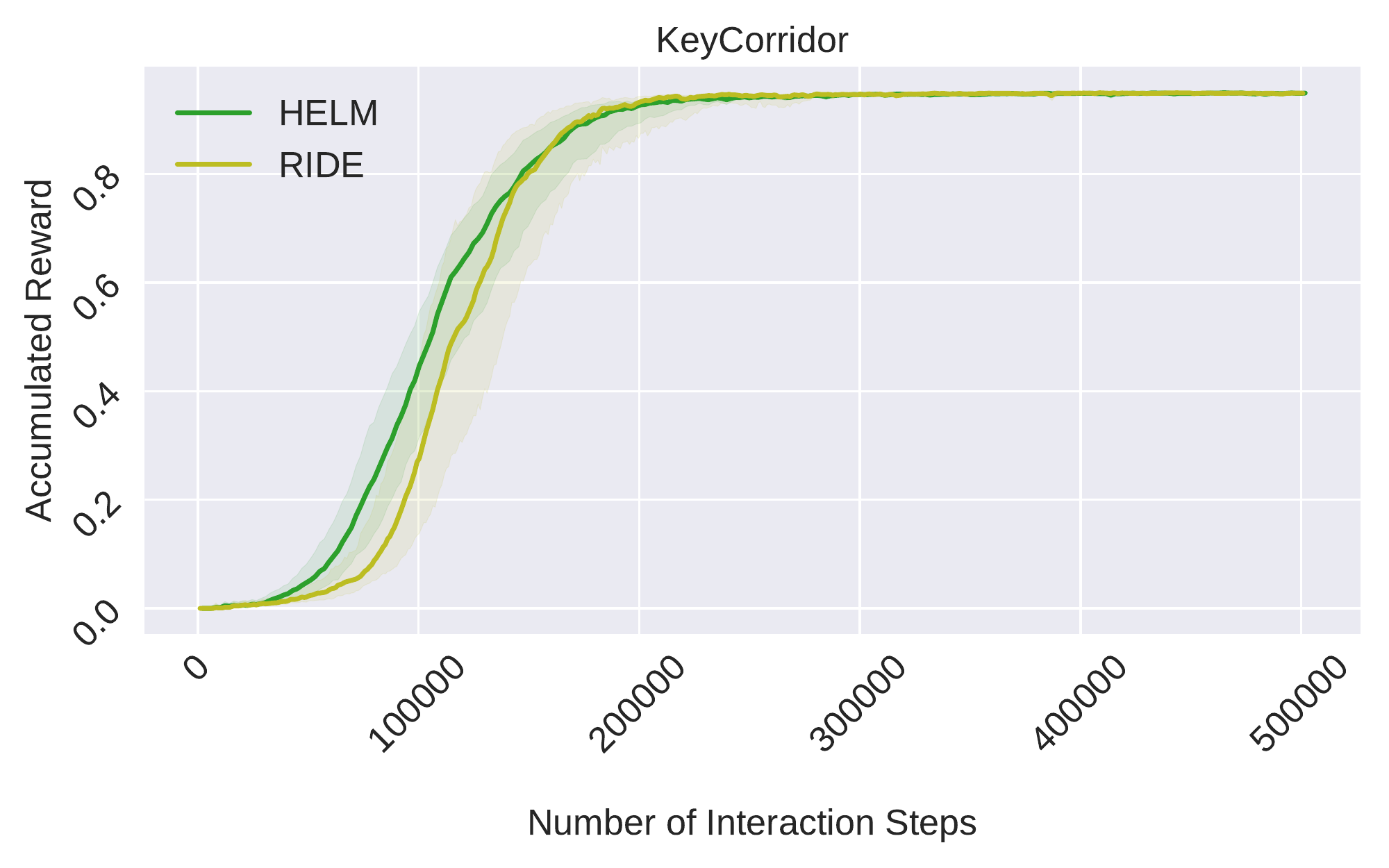}}
\caption{Comparison of \ourmethod~to a pretrained recurrent dynamics model (RIDE) used for history compression. IQM and 95\,\% bootstrapped CIs of return over last 100 episodes across 30 seeds are shown.}\label{fig:ablation_compression}
\end{center}
\vskip -0.2in
\end{figure}

\begin{figure}[p]
\vskip 0.2in
\begin{minipage}{.5\textwidth}
\centerline{\includegraphics[width=\textwidth]{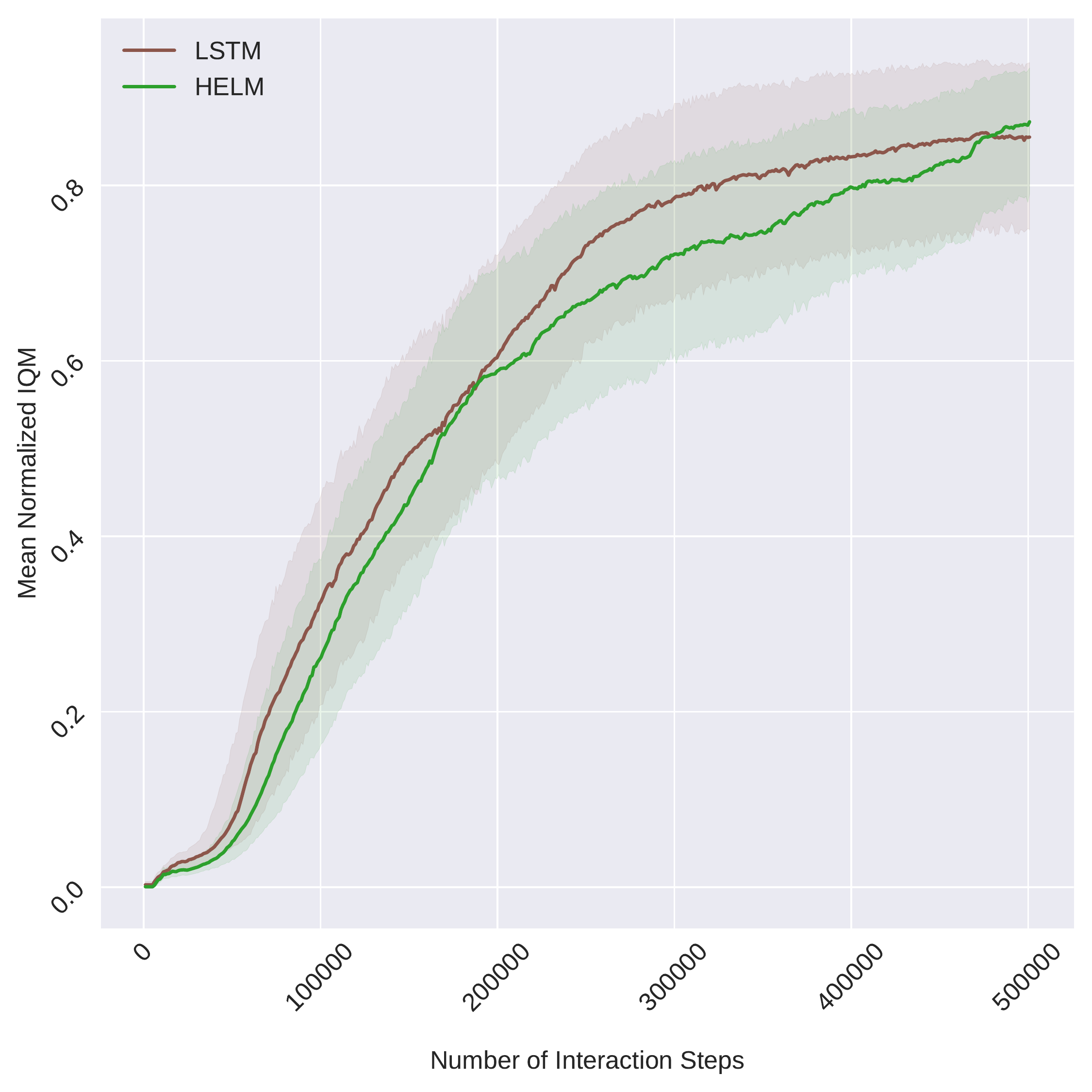}}
\end{minipage}
\begin{minipage}{.5\textwidth}
\centerline{\includegraphics[width=\textwidth]{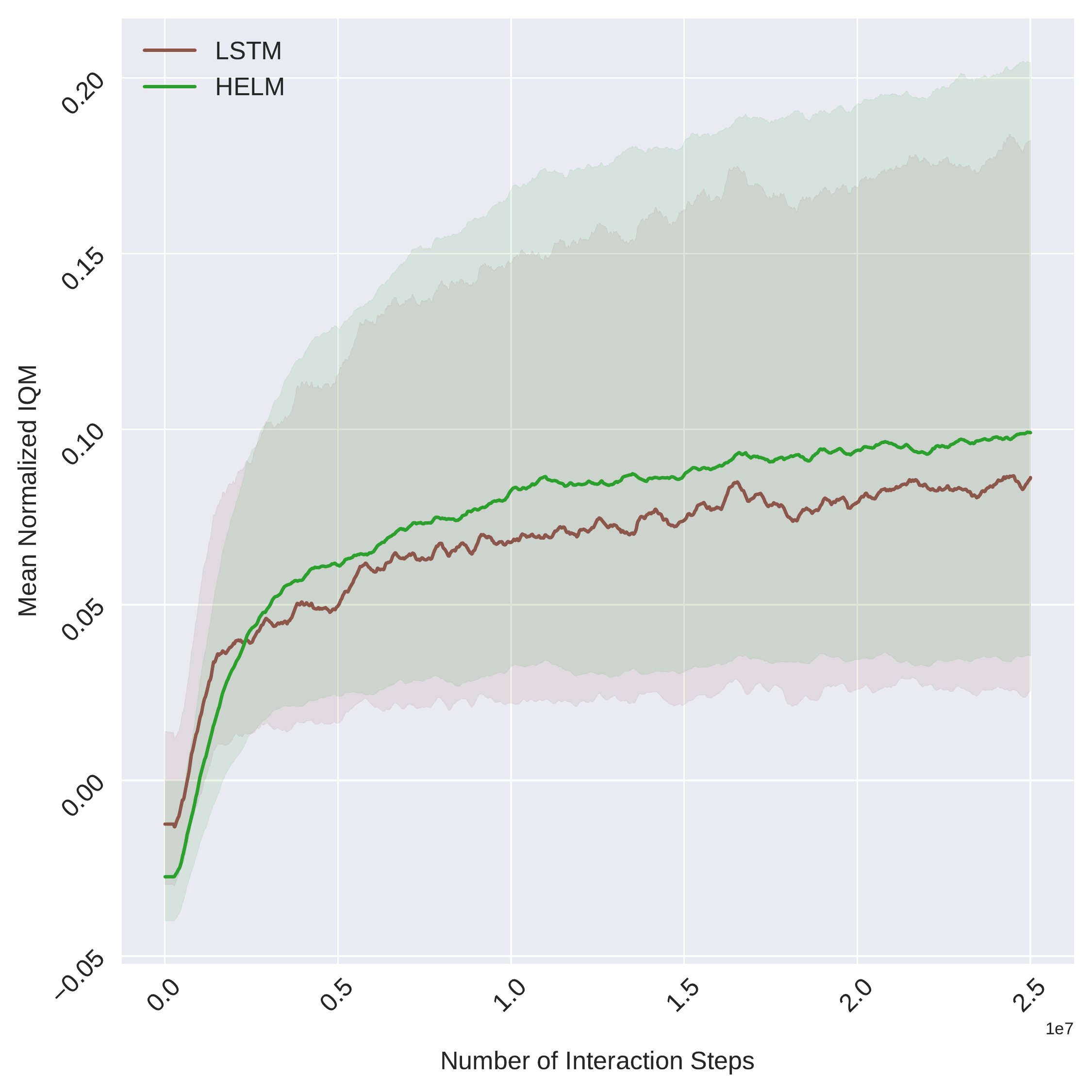}}
\end{minipage}
\caption{Comparison of \ourmethod~to an LSTM pretrained on natural language used for history compression. Mean IQM and 95\,\% bootstrapped CIs over all Minigrid and Procgen environments is shown.}
\label{fig:ablation_compression_lstm}
\vskip -0.2in
\end{figure}

While Transformer models excel when sufficient data is available it appears for low-resource settings like on-policy RL plenty of interaction steps with the environment are required for convergence.
In contrast, \ourmethod~is more sample efficient, stable, and exhibits lower variance than finetuning various components on the downstream task.
In line with findings of \citet{rothermel_dont_2021}, we observe that fully finetuning on the downstream task appears to be more stable than minimal finetuning of certain components.

Additionally, we conduct an ablation on exchanging TrXL with other models for history compression in RL. Particularly, we exchange TrXL with a pretrained LSTM language model and an LSTM pretrained on dynamics modeling on the \keycorr\ environment. 

We show performance of the following settings:
\begin{itemize}
    \item \textbf{\ourmethod}: Our proposed setting using \textit{\frozenpool}\ and TrXL. 
    \item \textbf{LSTM}: An LSTM language-model encoder \citet{merity_regularizing_2018} commonly used for transfer (\eg \citet{howard_universal_2018}), pretrained on WikiText-103 \citet{merity_pointer_2017}. We still use the \frozenpool~mechanism for mapping observations to the space of pretrained LSTM token embeddings.
    \item \textbf{RIDE}: An LSTM encoder trained on an intrinsic reward for exploration in procedurally generated environments proposed by \citet{raileanu_ride_2020}.
\end{itemize}

First, we compare \ourmethod\ to RIDE in \cref{fig:ablation_compression} on the \keycorr\ environment.
Remarkably, performance of \ourmethod\ is on-par with RIDE although RIDE is biased towards the dynamics of the \keycorr~environment.
In \cref{fig:ablation_compression_lstm} we compare \ourmethod\ against the pretrained LSTM encoder on natural language. 
Interestingly, LSTM reaches performance comparable with \ourmethod\ on both Minigrid and Procgen environments.
However, the final performance on both benchmark suites for \ourmethod\ is still slightly above the performance of LSTM.
These results provide evidence that the pretraining objective of language modeling is very well suited for history compression.
Also, architectural differences appear to play a role, as we consistently observe slightly better performance when relying on the Transformer architecture.
In the future we aim at conducting a more thorough comparison to different compression techniques such as the Compressive Transformer \citep{rae_compressive_2020}.

To highlight the importance of transfer from the language domain we conduct another set of experiments.
We train the following variants on the \keycorr\ environment:
\begin{itemize}
    \item \textbf{CNN-PPO+N}: Exchanging the output of TrXL with a random noise vector sampled from $\cN(0,1)$ to evaluate whether the output of TrXL carries more information than randomness.
    \item \textbf{\ourmethod-R}: Randomly initializing TrXL, since the effectiveness of random functions has been established in prior work on reservoir computing \citep{maass_model_2002,jaeger_echo_2001}.
    \item \textbf{\ourmethod+N}: Corrupting the output of TrXL with a random noise vector sampled from $\cN(0,1)$.
    \item \textbf{CNN-PPO+Pos}: Exchanging the output of TrXL with sinusoidal positional embeddings as used in \citet{devlin_bert_2019}, to determine whether the only additional information to effectively solve the task is the current position.
\end{itemize}

\begin{figure}[p]
\vskip 0.2in
\begin{center}
\centerline{\includegraphics[width=\textwidth]{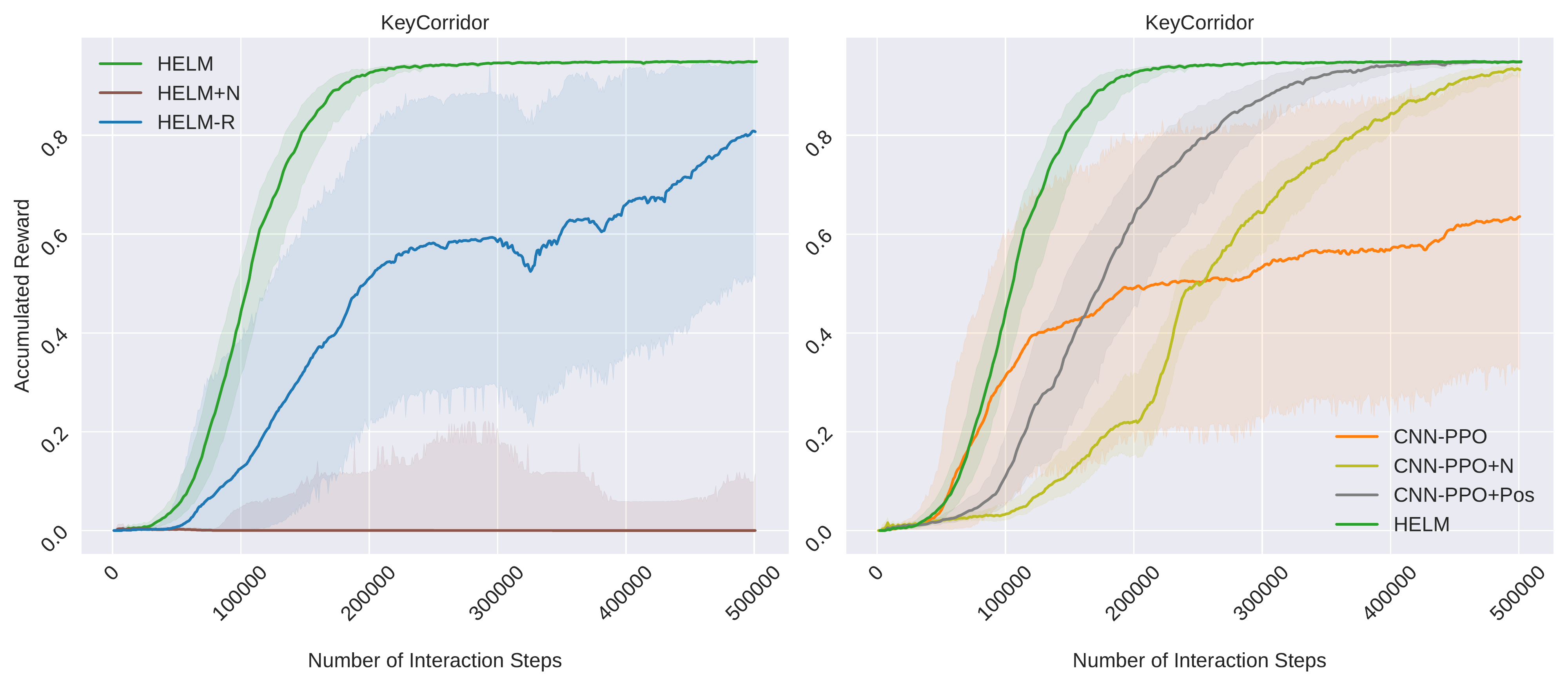}}
\caption{Noise injection experiments to highlight the transfer from the language domain. IQM and 95\,\% bootstrapped confidence interval of return over last 100 episodes across 30 seeds are shown.}\label{fig:noise_injections}
\end{center}
\vskip -0.2in
\end{figure}

We perform gridsearches as mentioned in \cref{app:hyperparam} for all aforementioned variants.
\cref{fig:noise_injections} right shows the comparison of \ourmethod\ to CNN-PPO, CNN-PPO+N, CNN-PPO+Pos.
We observe that exchanging the TrXL output with noise acts as regularization during the hyperparameter search of CNN-PPO, thus, to improved performance over CNN-PPO.
Augmenting CNN-PPO with positional information yields a further boost in performance.
Still, both variants are less sample efficient than \ourmethod, which indicates that the pretrained TrXL provides more useful information than positional information. 
\cref{fig:noise_injections} left shows the comparison of \ourmethod\ to \ourmethod-R, and \ourmethod+N.
Initializing TrXL at random results in a performance drop and increased variance, which illustrates the importance of the transfer from the language domain.
Furthermore, corrupting the output of TrXL with noise results diminishes performance.

\section{Notes on the Johnson-Lindenstrauss Lemma}
\label{app:jl_lemma}

The Johnson-Lindenstrauss lemma \cite{johnson_extensions_1984} states that the distances 
between points of a finite set are approximately preserved by a random projection with high probability. 
One can obtain this result by showing a similar property for a single point and then taking 
the union bound over all pairwise distances in the set. 
Since we want to apply the lemma to the observations in an environment, 
i.e., a possibly infinite set, we are interested in this more elementary version of the lemma. 
It can be found, e.g., in \citet{vershynin_high-dimensional_2018} lemma 5.3.2. 

In \cref{sec:methodology} we consider a random projection matrix
$\BP \in \dR^{m \times n}$ with $P_{ij}$ sampled i.i.d.\ from an appropriately scaled Gaussian distribution and an 
arbitrary but fixed vector $\Bd \in \dR^{n}$. 
We are interested in how much the random projection $\BP \Bd$ distorts the length of $\Bd$, 
i.e., we want to compare $\norm{\BP \Bd}$ to $\norm{\Bd}$. 

First, note that we can interchange the roles of $\BP$ and $\Bd$, 
i.e., we can choose $\Bd$ at random and leave $\BP$ fixed. 
To see this, let $\BR  \in \dR^{n \times n}$ be the rotation matrix that aligns $\Bd$
with the first axis and consider $\BP \BR^\top \BR \Bd$.
Due to rotation invariance of the Gaussian distribution, the distribution of $\BP \BR^\top$
is the same as that of $\BP$. The multiplication with $\BR \Bd$
corresponds to selecting the first column of $\BP \BR^\top$ and scaling it by $\norm{\BR \Bd}=\norm{\Bd}$. 
This is the same as choosing $\Bd$ at random with appropriate scaling and 
mapping it onto its first $m$ coordinates. 
Further, we may normalize $\Bd$ to unit length and scale the projection accordingly. 

Let $d_i \sim \cN(0, 1)$ and let $\Bz = \Bd / \norm{\Bd}$, i.e.,  
we choose $\Bz$ from the unit sphere uniformly at random. 
Further, let $\BQ \Bz$ be the projection of $\Bz$ onto its first $m$ coordinates. 
We have $1 = \dE[\norm{\Bz}_2^2] = \dE[\sum_{i=1}^n z_i^2] = n \dE[z_1^2]$ and, consequently, 
\begin{equation}
    \dE\left[\norm{\BQ \Bz}_2^2\right] = \dE\left[\sum_{i=1}^m z_i^2\right] = m \dE[z_1^2] = \frac m n.
\end{equation}
From lemma 2.2 \cite{dasgupta_elementary_2003} we conclude that for any $0 < \varepsilon < 1$
\begin{equation}
    \dP\left[\norm{\BQ \Bz}_2^2 \leq (1-\varepsilon) \frac m n \cup \norm{\BQ \Bz}_2^2 \geq (1+\varepsilon) \frac m n \right] \leq 2 \exp\left(-\frac{m(\varepsilon^2 / 2 - \varepsilon^3 / 3)}{2}\right).
\end{equation}
Therefore, the complementary event can bounded by 
\begin{equation}
    \dP\left[(1-\varepsilon) \leq \norm{\sqrt{\frac n m} \BQ \Bz}_2^2 \leq (1+\varepsilon) \right] \geq 1 - 2 \exp\left(-\frac{m(\varepsilon^2 / 2 - \varepsilon^3 / 3)}{2}\right).
\end{equation}
Equation \eqref{eqn:jl_approx} follows by multiplication with 
$\norm{\Bd}_2^2$ inside the probability and letting $\BP = \sqrt{n/m} \BQ$.

\section{Review of Modern Hopfield Networks}
\label{app:hopfield}

Hopfield networks are energy-based binary associative memories, which
popularized artificial neural networks in the 1980s
\citep{amari_learning_1972,hopfield_neural_1982,hopfield_neurons_1984}.
Associative memory networks have been designed to store and retrieve samples. 
Their storage capacity can be considerably increased 
by polynomial terms in the energy function
\citep{chen_high_1987,psaltis_nonlinear_1986,baldi_number_1987,gardner_multiconnected_1987,abbott_storage_1987,horn_capacities_1988,caputo_storage_2002,krotov_dense_2016}.

In contrast to these binary memory networks, continuous MHNs utilize an associative memory with exponential storage capacity. 
These MHNs for deep learning architectures have 
an energy function with continuous states and can
retrieve samples with only one update \citep{ramsauer_hopfield_2021}. 

We assume a set of patterns (or embeddings) $\{\Be_1,\ldots,\Be_k\} \subset \dR^m$
that are stacked as row vectors to form
the matrix $\BE = (\Be_1, \dots, \Be_k)^\top$ and a 
state pattern (query) $\Bxi \in \dR^m$ that represents the current state. 
In the \textit{\frozenpool} mechanism we obtain the state pattern by 
a random but fixed mapping $\Bxi = \BP \Bo_t$.
The largest norm of a stored pattern is
$M = \max_{i} \norm{\Be_i}_2$.
Continuous MHNs with state $\Bxi$
have the energy
\begin{align}
L = -\beta^{-1} \log \left( \sum_{i=1}^m
\exp(\beta \Be_i^\top \Bxi) \right) +  \beta^{-1} \log k  +  
\frac{1}{2} \Bxi^\top \Bxi + \frac{1}{2} M^2.
\end{align}
For energy $L$ and state $\Bxi$, the update rule 
\begin{align}
\label{eq:Amain_iterate}
\Bxi \gets f(\Bxi) = \BE^\top \sigma(\beta \BE \Bxi)
\end{align}
has been proven to converge globally  
to stationary points of the energy $L$, 
which are almost always local minima 
\citep{ramsauer_hopfield_2021}.
The update rule \eqref{eq:Amain_iterate}
is also the formula of the well-known Transformer attention mechanism
\citep{vaswani_attention_2017}. Therefore, Hopfield retrieval and
Transformer attention coincide.

The {\em separation} $\Delta_i$  of a 
pattern $\Be_i$ is defined as its minimal dot product difference to any of the other 
patterns:
$\Delta_i = \min_{j,j \neq i} ( \Be_i^\top \Be_i - \Be_i^\top \Be_j )$. 
A pattern is {\em well-separated} from the data if $
 \Delta_i  \geq \frac{2}{\beta k} + \frac{1}{\beta} \log \left( 2 (k-1)  k  \beta  M^2 \right)$.
If the patterns $\Be_i$ are well separated, the update rule~\eqref{eq:Amain_iterate}
converges to a fixed point close to a stored pattern.
If some patterns are similar to one another and, therefore, not well separated, 
the update rule~\eqref{eq:Amain_iterate} converges to 
a fixed point close to the mean of the similar patterns. 
This fixed point is a {\em metastable state} of the energy function
and averages over similar patterns.

The \textit{\frozenpool} mechanism can be viewed as a retrieval with one update. 
The next theorem states that the update rule \eqref{eq:Amain_iterate} typically converges after
one update if the patterns are well separated. Furthermore, it states
that the retrieval error is 
exponentially small in the separation $\Delta_i$.
\begin{theorem}[Modern Hopfield Networks: Retrieval with One Update]
\label{th:AoneUpdate}
With query $\Bxi$, after one update the distance of the new point $f(\Bxi)$
to the fixed point $\Be_i^*$ is exponentially small in the separation $\Delta_i$.
The precise bounds using the Jacobian $J = \frac{\partial
  f(\Bxi)}{\partial \Bxi}$ and its value $\bar J$ in the mean value
theorem are:
\begin{align}
  \norm{f(\Bxi) - \Be_i^*}
  &\leq  \norm{\bar J}_2 \norm{\Bxi - \Be_i^*}  , \\
  \norm{\bar J}_2  &\leq 2 \beta k M^2 (k-1) \exp(- \beta (\Delta_i - 2  \max \{ \norm{\Bxi - \Be_i} , \norm{\Be_i^* - \Be_i} \} M) ).
\end{align}
For given $\varepsilon$ and 
sufficient large $\Delta_i$, we have $\norm{f(\Bxi) - \Be_i^*} < \varepsilon$,
that is, retrieval with one update.
The retrieval error $\norm{f(\Bxi) - \Be_i}$ of pattern $\Be_i$
is bounded by
\begin{align}
   \norm{f(\Bxi) - \Be_i} &\leq 2 (k-1) \exp(- \beta (\Delta_i - 2 \max \{ \norm{\Bxi - \Be_i} , \norm{\Be_i^* - \Be_i} \} 
   M) )  M .
 \end{align}
\end{theorem}
For a proof see \citet{ramsauer_hopfield_2021}.

The main requirement of MHNs to
be suited for association of different modalities
is that they can store and retrieve enough embeddings of the target embedding space.
We want to store a potentially large set of embeddings, since TrXL was pretrained with over 260K tokens.
We first define what we mean by storing and retrieving patterns
from a MHN.
\begin{definition}[Pattern Stored and Retrieved]
We assume that around every pattern $\Be_i$ a sphere $S_i$ is given.
We say $\Be_i$ {\em is stored} if there is a single fixed point $\Be_i^* \in S_i$ to
which all points $\Bxi \in S_i$ converge,
and  $S_i \cap S_j = \emptyset$ for $i \neq j$.
We say $\Be_i$ {\em is retrieved} for a given $\varepsilon$ if 
update rule \eqref{eq:Amain_iterate} gives
a point $\tilde{\Bx}_i$ that is at least 
$\varepsilon$-close to the single fixed point $\Be_i^* \in S_i$. 
The retrieval error is $\norm{\tilde{\Bx}_i - \Be_i}$.
\end{definition}

As with classical Hopfield networks, we consider patterns on the sphere, 
i.e., patterns with a fixed norm. 
For randomly chosen patterns, the number of patterns that can be stored
is exponential in the dimension $m$ of the space of the patterns ($\Be_i \in \dR^m$).
\begin{theorem}[Modern Hopfield Networks: Exponential Storage Capacity]
\label{th:Astorage}
We assume a failure probability $0 < p \leq 1$ and randomly chosen patterns 
on the sphere with radius $M:=K \sqrt{m-1}$. 
We define $a := \frac{2}{m-1}  (1 + \ln(2 \beta K^2 p (m-1)))$, 
$b := \frac{2  K^2  \beta}{5}$,
and $c:= \frac{b}{W_0(\exp(a + \ln(b))}$,
where $W_0$ is the upper branch of the Lambert $W$ function \dlmf{4.13},
and ensure $c \geq \left( \frac{2}{ \sqrt{p}}\right)^{\frac{4}{m-1}}$.
Then with probability $1-p$, the number of random patterns 
that can be stored is  
\begin{align} 
 \label{eq:ACapacityM}
    k \ &\geq \ \sqrt{p} \ c^{\frac{m-1}{4}}  \ .
\end{align}
Therefore it is proven for $c \geq 3.1546$ with
$\beta=1$, $K=3$, $m = 20$ and $p = 0.001$ ($a + \ln(b)>1.27$)
and proven for $c\geq 1.3718$ with $\beta = 1$, $K = 1$, $m = 75$, and $p=0.001$
($a + \ln(b)<-0.94$).
\end{theorem}
For a proof see \citet{ramsauer_hopfield_2021}.

\begin{figure}[p]
\vskip 0.2in
\begin{center}
\centerline{\includegraphics[width=.9\textwidth]{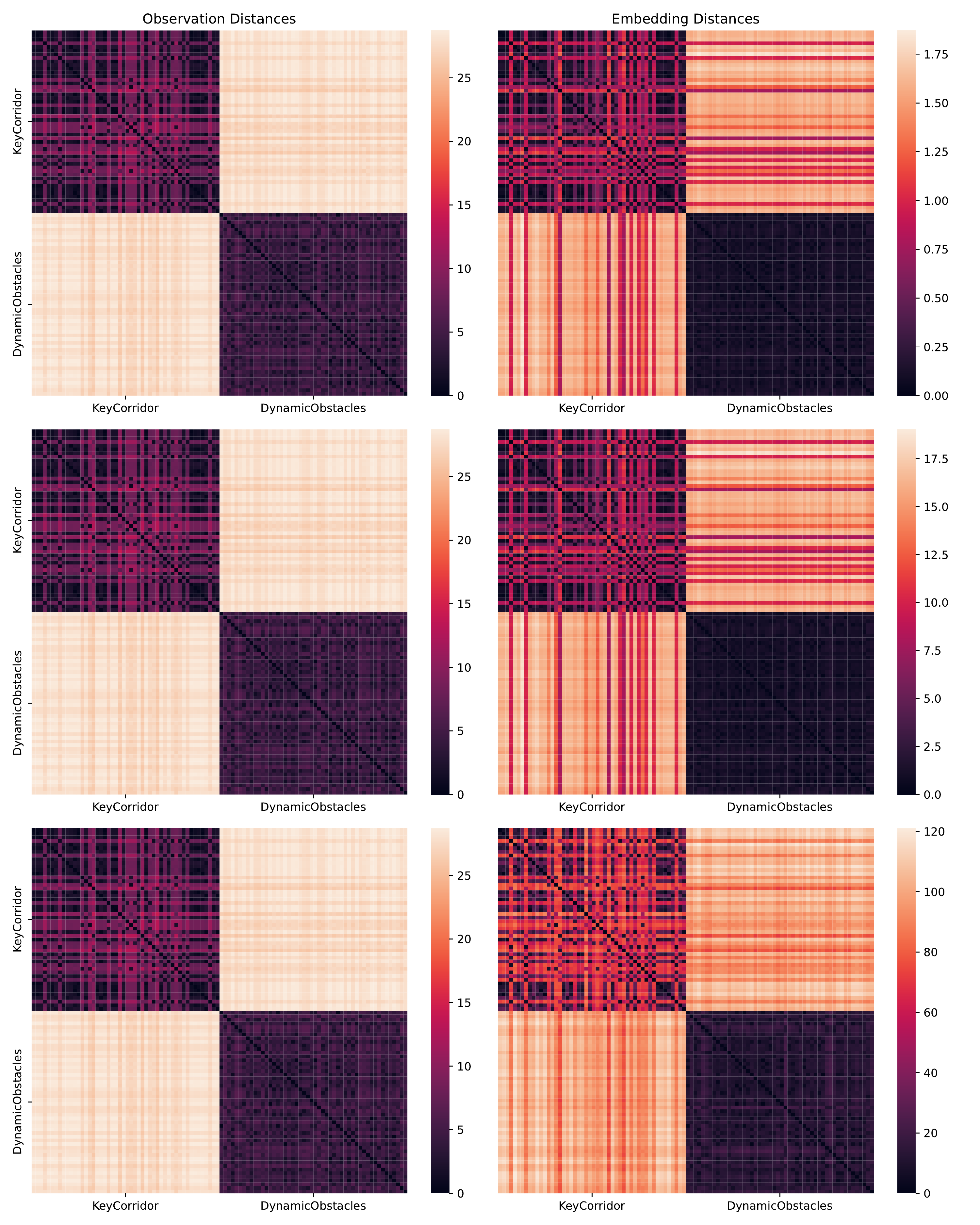}}
\caption{Distance matrices for observations of the \keycorr\ and \dynobst\ environments for different values of $\beta \in \{1, 10, 100 \}$. Image observations are very well separated in pixel space (left column). After mapping the observations to the PLT input space the distances are well separated (right column). Increasing $\beta = 1$ (top) to $\beta = 10$ (middle) the distances in the embedding space are also scaled up by an order of magnitude. Further increasing the temperature parameter to $\beta = 100$ results in enhanced distances across environments, while inter-environment distances are also enhanced. Thus, the parameter $\beta$ can avoid representation collapse in the token embedding space. }\label{fig:dist_mats}
\end{center}
\vskip -0.2in
\end{figure}

\begin{figure}[p]
\vskip 0.2in
\begin{center}
\centerline{\includegraphics[width=\textwidth]{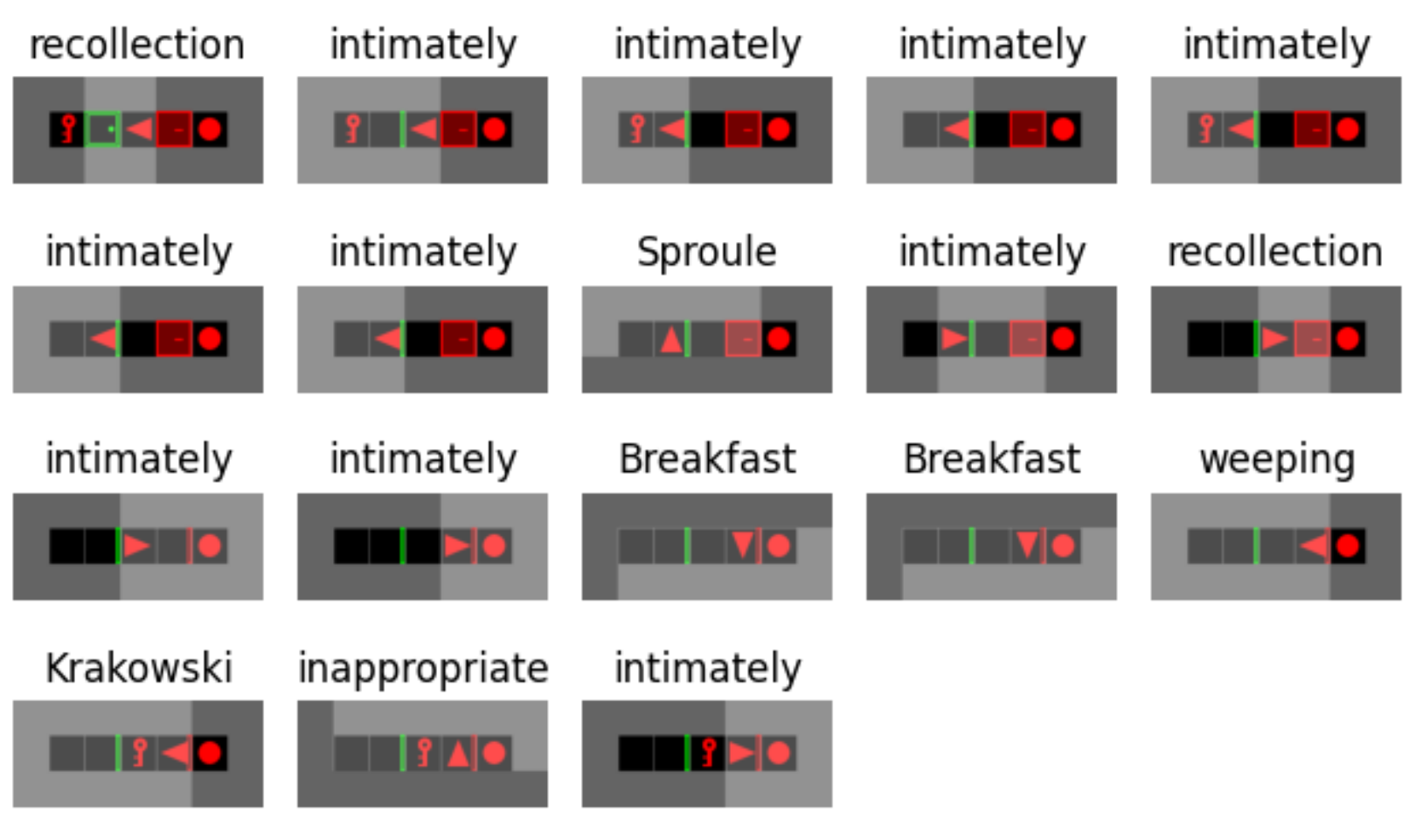}}
\caption{Sample episode of a trained policy with token annotations corresponding to the closest token in the PLT embedding space. The episode starts at left top and ends at right bottom. The token annotations are not meaningful since the mapping to the token space is initialized at random. Still, similar states in the pixel space map to the same token, e.g., ''recollection`` when the agent faces a door in front of it.}\label{fig:sample_episode}
\end{center}
\vskip -0.2in
\end{figure}

\end{document}